\documentclass[conference]{IEEEtran}

\usepackage{balance}
\usepackage{rotating}
\usepackage{amsmath,amsfonts}
\usepackage{graphicx}
\usepackage{textcomp}
\usepackage{xcolor}
\def\BibTeX{{\rm B\kern-.05em{\sc i\kern-.025em b}\kern-.08em
    T\kern-.1667em\lower.7ex\hbox{E}\kern-.125emX}}
    
\usepackage[official]{eurosym}
\usepackage{placeins}
\usepackage{graphicx}
\usepackage{textcomp}
\usepackage{color}
\usepackage[utf8]{inputenc}
\usepackage{dblfloatfix}
\usepackage{float}
\usepackage{arcs}
\usepackage{booktabs}
\usepackage{color,soul}
\usepackage{xcolor}
\usepackage{latexsym}
\usepackage{psfrag}
\usepackage{tikz}
\usepackage{color, colortbl}
\usepackage{algorithmicx}
\usepackage{algcompatible}
\usepackage[font=small,labelfont=bf]{caption}
\usepackage[tight]{subfigure}

\usepackage{pifont}
\usetikzlibrary{arrows,shapes}
\usepackage{multirow}
\usepackage{url}
\usepackage{xspace}
\usepackage{pdfpages}
\usepackage{tabularx}
\usepackage{adjustbox}
\usepackage{flushend}
\usepackage{balance}
\newtheorem{definition}{Definition}

\newcommand{\eg}{{\it e.g., }}
\newcommand{\ie}{{\it i.e., }}

\newcommand{\sol} {{ReMI}\xspace}
\def\BibTeX{{\rm B\kern-.05em{\sc i\kern-.025em b}\kern-.08em
    T\kern-.1667em\lower.7ex\hbox{E}\kern-.125emX}}
\makeatletter
\def\@copyrightspace{\relax}
\makeatother

\begin{document}

\title{Silver Linings in the Shadows: Harnessing Membership Inference for Machine Unlearning}
%


 \author{
 \IEEEauthorblockN{Nexhi Sula}
 \IEEEauthorblockA{Saint Louis University\\
 St. Louis, MO, US\\
 nexhi.sula@slu.edu}
 \and
 \IEEEauthorblockN{Abhinav Kumar}
 \IEEEauthorblockA{Saint Louis University\\
 St. Louis, MO, US\\
 abhinav.kumar@slu.edu}
 \and
 \IEEEauthorblockN{Jie Hou}
 \IEEEauthorblockA{Saint Louis University\\
 St. Louis, KS, US\\
 jie.hou@slu.edu}
 \and
 \IEEEauthorblockN{Han Wang}
 \IEEEauthorblockA{University of Kansas\\
 Lawrence, MO, US\\
 han.wang@ku.edu}
 \and
 \IEEEauthorblockN{Reza Tourani}
 \IEEEauthorblockA{Saint Louis University\\
 St. Louis, MO, US\\
 reza.tourani@slu.edu}
 }


%
%
\maketitle
%
\begin{abstract}

With the continued advancement and widespread adoption of machine learning (ML) models across various domains, ensuring user privacy and data security has become a paramount concern. 
In compliance with data privacy regulations, such as GDPR, a secure machine learning framework should not only grant users the right to request the removal of their contributed data 
used for model training but also facilitates the elimination of sensitive data fingerprints within machine learning models to mitigate potential attack -- a process referred to as machine unlearning. 

In this study, we present a novel unlearning mechanism designed to effectively remove the impact of specific data samples from a neural network while considering the performance of the unlearned model on the primary task. In achieving this goal, we crafted a novel loss function tailored to eliminate privacy-sensitive information from weights and activation values of the target model by combining target classification loss and membership inference loss. Our adaptable framework can easily incorporate various privacy leakage approximation mechanisms to guide the unlearning process. We provide empirical evidence of the effectiveness of our unlearning approach with a theoretical upper-bound analysis through a membership inference mechanism as a proof of concept. Our results showcase the superior performance of our approach in terms of unlearning efficacy and latency as well as the fidelity of the primary task, across four datasets and four deep learning architectures.
\end{abstract}
%

%
\section{Introduction}
\label{sec:introduction}
\vspace{0.08in}
\hfill
\parbox{0.88\columnwidth}{
\small
\textit{``The data subject shall have the right to obtain from the controller the erasure of personal data concerning him or her without undue delay and the controller shall have the obligation to erase personal data without undue delay ...''}
\hrule
\raggedleft
\vspace{0.02in}
\small{Article 17 GDPR}}
\vspace{0.15in}

In the era of data-driven technological innovation, 
the need for data collection and processing intensifies with the widespread adoption and increasing prominence of Machine Learning (ML) applications, such as Google Gboard, Watson for Oncology, or OpenAI ChatGPT, in our daily lives. As the collection of users' private information 
grows at an unprecedented rate, it is of paramount importance to ensure the security of this data and uphold the requirements of user privacy.
To enhance and standardize data protection and user privacy, several legislations have been enacted, including the European Union's General Data Protection Regulation (GDPR)~\cite{gdpr,c6} and the California Consumer Privacy Act (CCPA)~\cite{c5}. 
In recent years, GDPR violations have resulted in fines for various companies, \eg in 2022, the Spanish data protection authority imposed a fine of \euro {10} million on Google LLC for violating \emph{Articles 6 and 17} of GDPR~\cite{dataguidance-google-fine}.
%

Unlike erasing data at rest, removing data samples or their influence from a trained neural network is extremely complex. This is in part due to neural networks' memorization tendency, especially when dealing with complex models, limited training data, and the presence of long-tailed distributions in image data~\cite{ArpJasBal17,CarLiuErl19,FelZha20}. 
When a model memorizes specific data samples, including those containing sensitive or private information, it becomes susceptible to privacy attacks, such as membership inference attacks (MIA)~\cite{TruLiuGur19,NasShoHou19,ShoStrSon17}. 
Even well-generalized models have shown to be vulnerable to such attacks despite using regularization protection~\cite{LonBinWan18,BalSheHit22}. The growing concern for data privacy and the need to remove personal data instances have fueled the field of \emph{machine unlearning}.
%

The most evident and provable machine unlearning solution is \emph{naïve retraining} -- training a newly initialized model after excluding the sample that should be forgotten. While viable, retraining is computationally expensive and only possible if the original data is available~\cite{c2}, which makes it unsuitable for most of the modern machine learning models.
Exact unlearning is a class of unlearning methods that aims to alter the training algorithms to generate unlearned models identical to retraining while reducing the cost~\cite{c1,c16,yan2022arcane,c15,c25,c26}. The most prominent exact unlearning solution, SISA~\cite{c1}, relies on sharding and slicing the data to train multiple sub-models, which could be retrained individually. Alternatively, batch updates can be logged during training, allowing the removal of individual batch updates~\cite{c15}, though significantly increasing the storage and deployment cost.
To satisfy cost constraints in modern learning paradigms, there is a relaxation of the requirement for the unlearned model to be identical to retraining. Thus, leading to approximate unlearning algorithms~\cite{c19,c22,guochugol19,GolAdiAch20,ThuDezCha22,c33}. These mechanisms rely on utilizing the information provided by the Hessian~\cite{c19,c21,guochugol19} or use architectural modifications~\cite{c33} to get a final learned representation that approximately resembles retraining. More so, they require access to either the training data or certain training process information, such as the Fisher Information matrix, which may not be accessible during unlearning.
%

While these methods partially fulfill the requirements of machine unlearning, they also introduce certain strong assumptions regarding training process auditing, storage requirements, or modification of the training paradigm. This has motivated us to design an unlearning framework, aiming to provide high unlearning efficacy and efficiency while offering flexibility in sample forgetting, even if the training data is not accessible. Moreover, we seek a solution that is retrospectively applicable to formerly trained models, irrespective of the training process. 
%
%
%

To achieve the ``right to be forgotten'' and completely remove requested samples (\ie forgetting samples) along with their influence from the model, we propose {\em \sol Unlearning}, which stands for {\em Reverse Membership Inference Unlearning}. At its core, \sol aims to scrub the unique and identifiable features of the forget samples from the model, ensuring that such information is not retained in the unlearned model and therefore cannot be inadvertently leaked or extracted. 
These key features 
include model's posterior vector, predicted label, loss, gradients, and intermediate activation values. These features, often exploited in orchestrating various privacy attacks~\cite{ShoStrSon17,LonBinWan18,NasShoHou19,TruLiuGur19}, play a crucial role in guiding the unlearning process. 
In \sol, we establish the notion of a {\it privacy approximation function} as the one that extracts sensitive training information from the model. This function can be realized as a neural network trained on the key features, extracted from the model given a sample set with an identical distribution to the training dataset. In essence, it is a more generalized adaptation of membership and attribute inference attack models, which we use to guide the unlearning process. We further introduce an alternative approach -- {\it Membership Fingerprinting} (MF) model -- to aid the unlearning process. The {\it Membership Fingerprinting} model eliminates the need for training an additional shadow model from scratch on new datasets, making it computationally more efficient and practical. 

We devise a customized unlearning loss function that combines the 
target model's loss with 
the privacy approximation function's loss. The 
first component retains the unlearned model's fidelity for the primary task, while the latter loss component evaluates the residual influence of forgetting samples and minimizes the potential extraction of sensitive features by the privacy approximation function. 
Utilizing this loss function in \sol's design allows for a versatile range of removal options, from eliminating a single sample to removing multiple samples from one class or multiple classes, which has shown to be more challenging that removing an entire class or a subset of a single class~\cite{CheQiZih24}. 
%

We evaluate the performance of our unlearning mechanism using FMNIST, UTKFace, STL10, and CIFAR-10 datasets, across four neural network architectures. To demonstrate \sol efficacy in unlearning, we use MIA implementation in~\cite{liu2022ml,nicolae2018adversarial} and epistemic uncertainty~\cite{BecLie22} to quantify the amount of information the unlearned model leaks when given the forgotten samples. While using MIA accuracy to assess unlearning effectiveness may be controversial, prior studies have indeed adopted this approach~\cite{c19}. We also evaluate the performance of the unlearned model on the primary classification task for the remaining data to ensure it retains high fidelity, \ie comparable to the original target model. Our empirical findings, obtained with various neural network architectures, affirm the effectiveness of our tailored unlearning loss functions.

\textbf{Contributions:} In summary, we study the problem of removing samples of training data and eliminating their influence from neural networks in practicing the right to be forgotten. The contributions of this paper are: 
\begin{itemize}
\item We introduce \sol, an unlearning mechanism that effectively expunges selected training samples from one or multiple classes and their impact from a pre-trained neural network. \sol neither requires prior knowledge of the training process nor stores any parameters during training. 
\item In designing \sol, we propose a novel unlearning loss function to minimize the information leakage of the unlearned model. At its core, \sol uses privacy-sensitive features of samples, often used in inference attacks, to measure the amount of information that the original leaks and guide the unlearning process.
\item When driving the customized unlearning loss function, we provide a new definition quantifying the indistinguishability of forgetting samples during the unlearning process. Furthermore, we establish a theoretical upper bound for this indistinguishability level. Incorporating this upper bound into our customized unlearning loss function enables us to minimize the risk of information leakage from forgotten samples while ensuring the preservation of utility for classification tasks.
\item We conducted extensive experiments to validate the performance of \sol on multiple architectures and datasets. Our proposed method was compared with two benchmarks: Fisher unlearning~\cite{c19} and naïve retraining.  In our evaluation, we analyze the advantages and shortcomings of each approach.
\end{itemize}

\section{Related Work}
\label{sec:relatedwork}
The task of removing selected data from a trained model has attracted researchers due to its potential ability to improve model security, privacy, and usability~\cite{c3}. To realize this task, researchers have proposed several different machine unlearning frameworks and approaches. This section provides an overview of the existing techniques and their limitations to motivate the development of our design. The existing techniques can broadly be grouped into two categories: Exact Unlearning and Approximate Unlearning.

{\bf Exact Unlearning} or perfect unlearning algorithms formulate the goal of an unlearning mechanism as generating a posterior or weight distribution that is identical to the distribution of the model trained on the original training dataset, excluding the samples being forgotten. Exact unlearning methods either require significant altering of the training paradigm, like dividing the dataset into shards and making sub-models for each shard of dataset~\cite{c1,c28,CheZhaWan22}, or performing heavy auditing of the training process by storing every single batch update during the training process~\cite{c26}. While these methods provide provable unlearning guarantees, they incur significant overhead during both training and deployment periods, and cannot be retrospectively applied to already trained models. 

{\bf Approximate Unlearning} algorithms attempt to satisfy the constraints on training and deployment cost while making relatively fewer changes in the learning algorithm or pipeline. The relaxed constraints allow the unlearned model to be a close approximation of a model trained on the training dataset excluding the forgotten samples. Depending on the desired privacy setting, various approximate unlearning algorithms have been proposed. These methods may involve adjusting model weights using techniques like Hessian approximation~\cite{c19,c22,guochugol19}, leveraging neural tangent kernels~\cite{GolAdiAch20}, or modifying model architecture~\cite{c33}.  Alternatively, some approaches focus on auditing the training process and unrolling updates to retain information about the forgotten samples~\cite{ThuDezCha22}. However, the majority of these approaches require access to the training data during the unlearning process~\cite{ThuDezCha22}, or they depend on information like Fisher Information Matrix (FIM) approximations~\cite{c19,c25}, which must be computed during training. These requirements pose strong assumptions that limit the practicality of these unlearning mechanisms. Unlearning methods proposed without these assumptions~\cite{ChuTarMan23} have poor performance due to their reliance on error maximization-minimization noise to generate an unlearned model, leading to significant degradation in performance due to ad-hoc noise generation process~\cite{c3}.
Some unlearning solutions are specific to a given learning algorithm, such as Random Forests~\cite{brolow21}, Naive Bayes~\cite{NguLowJai20}, and linear models~\cite{c22}.

We design \sol to address cost constraints that are often encountered in practical unlearning scenarios. \sol is compatible with previously trained models and does not rely on the strong assumption of training data availability. It can be applied to a wide range of training algorithms and supports the unlearning of samples from multiple classes while minimizing the impact on model performance. \sol does not necessitate prior knowledge of the training process, eliminating any associated storage costs.
\section{Preliminaries}
\label{sec:preliminaries}
In this section, we briefly reason our choice of sample unlearning and then formally define the machine unlearning problem. We then elaborate on the privacy risks of neural networks, sensitive features that are commonly used in privacy attacks, and their role in our mechanism. 
%
%
Machine unlearning can be broadly categorized into three primary types: sample unlearning, feature unlearning, and class unlearning. Sample unlearning entails removing specific data points or instances from the model's training dataset, allowing the model to forget the influence of certain examples it has previously learned~\cite{c31,c22,c24,c32}. Class unlearning is one generalization of sample unlearning in which the samples to unlearn constitute an entire class. Thus, enabling the model to update its predictions and associations with specific classes, effectively undoing previous classifications~\cite{c2,c33,c34}. Feature unlearning, on the other hand, focuses on modifying or suppressing the influence of particular attributes or characteristics in the model's input data, enabling the model to forget correlations or patterns associated with those features~\cite{c2,c35}.
Sample unlearning is the foundation and more challenging task~\cite{CheQiZih24}, which can be extended into class and feature unlearning. As such, we specifically designed the machine unlearning approach for sample unlearning.

\subsection{Formalizing Machine Unlearning}
\label{sec:formalization}
We formalize the unlearning problem in the context of the Machine-Learning-as-a-Service paradigm between a machine learning service provider and a group of clients, $\mathcal{C}=\{c_1,c_2, \cdots, c_m\}$. Generically, the service provider can be envisioned as an entity with sufficient resources to train a (potentially proprietary) machine learning model, $\mathcal{M}$. To initiate the training and testing processes, the service provider collects diverse data samples from the client group and curates a dataset, 
$D = \{(x^{j}_i, y^{j}_i) \mid i \in \{1, \cdots, N\}, j \in \{1, \cdots, m\}, y \in \{1, \cdots, K\}\}$. 
In this dataset, $x^{j}_i$ represents the features for the {\it i-th} sample collected from client j, while $y^{j}_i$ denotes the label (\ie class) associated with that sample, chosen from a set of K possible classes. The learning model $\mathcal{M}$ is derived by optimizing the loss function to find the optimal weights that satisfy the objective function 
$w = \operatorname*{argmin}_w{\mathcal{L}_{D}(y, \mathcal{M}(x))}$.

As privacy laws, such as GDPR, state that data owners can request the timely removal of their identifiable data from the service provider, machine unlearning aims to remove the private information of the data owners (\ie clients) from a trained machine learning model, safeguarding it from potential attackers. More specifically, upon client $c$'s request for the removal of their private data, $D_f^c \subset D$, the service provider should transform model $\mathcal{M}$ (which has been trained on $D$) to $\mathcal{M}_u^c$, where the latter signifies the model that is not influenced by $D_f^c$. We consider $\mathcal{M}_u^c$ as the unlearned model. Note that $D_f^c$ can be either a single data sample or a set of samples belonging to client group $c$. To simplify the notation, in the rest of the paper, we omit using the client's identifier and represent forgetting data as $D_f$ and unlearned model as $\mathcal{M}_u$.
%

Given the difficulty of precisely measuring the influence of $D_f$ on $\mathcal{M}$, inverting it to a perfect unlearned model is a fairly challenging task, especially without prior knowledge of $D_f$ during training or the storage of detailed training process parameters. The ideal approach for unlearning $D_f$ is to train a new neural network (with similar hyperparameters as $\mathcal{M}$) using the remaining data $D_r = D \backslash D_f$, resulting in a trained model $\mathcal{M}_r$, that has never been impacted by $D_f$ -- dubbed as retraining. In the existing literature, retraining is considered the gold standard unlearning mechanism and the baseline for comparison~\cite{c26,c19,BecLie22}. While retraining can achieve optimal performance with respect to the client privacy, it is a highly inefficient solution, particularly when $|D \backslash D_f| \gg |D_f|$ or the model training is costly due to the model complexity. 


\subsection{Privacy Risks in Deep Learning}
\label{sec:privacy}
%
%
We define a function, denoted as $I(w)$, for extracting knowledge from a target model $\mathcal{M}$ with the objective of inferring private information related to the original training data owned by clients. The knowledge generated by the function $I(w)$ is commonly referred to as {\it attack features}. In the context of a black-box attack setting, the function $I(w)$ generates a collection of model outputs associated with each data point. This collection encompasses several key elements, including the output probabilities denoted as $\hat{p} = \mathcal{M}(x)$, the predicted class label ($\hat{y}$), determined as $\operatorname*{argmax}_c P(y=c|x)$, and the predictive loss, represented as $\mathcal{L}(\hat{y},y)$. In the white-box attack setting, the function $I(w)$ extracts additional information from the model, which includes gradients $\nabla \mathcal{L} = [\dfrac{\partial \mathcal{L}}{\partial w_1}, \dfrac{\partial \mathcal{L}}{\partial w_2},\ldots,\dfrac{\partial \mathcal{L}}{\partial w_m}]$ and intermediate activation values from layer $h$ of $\mathcal{M}$, denoted as $[a]_h$. We generalize this notation to $[a]_\mathcal{M} = \{[a]_1, [a]_2, \cdots, [a]_H\}$ for all the $H$ layers of model $\mathcal{M}$. 
As such, function $I(w)$, which acts as the generator of attack features plays a crucial role in various privacy attack algorithms.
These attack algorithms encompass a range of techniques, including membership inference attacks~\cite{TruLiuGur19,NasShoHou19}, property inference attacks~\cite{WanSonZha19,XuLi20}, data reconstruction attacks~\cite{FreJhaRis15,HitAtePer17,WanSonZha19}, and poisoning-based inference attacks~\cite{TraShoSan22,ChaAbaOpr23}. 

To successfully infer the private information of the training data, it's essential to have a well-trained machine learning attack model, which can effectively leverage the model information contained within the attack feature sets $I(w) \subset \{\hat{p}, \hat{y}, \mathcal{L}(\hat{y},y), \nabla \mathcal{L}, [a]_\mathcal{M}\}$ to retrieve the private information associated with the forgotten data.
We acknowledge that $I(w)$ generates a non-inclusive list of features; there are other privacy attack techniques, which use other features, such as loss trajectories~\cite{LiuZhaBac22}. Nonetheless, the attack features we mentioned are among the most common ones.

In the unlearning scenario, we propose a weight refinement algorithm to eliminate sensitive information from the attack features generated by $I(w)$, which are initially encoded by the pre-trained weights within model $\mathcal{M}$. In particular, we define a successful unlearning algorithm as one that provides an updated model with weights that meet the following objective function:
\setlength{\abovedisplayskip}{2pt}%
\setlength{\belowdisplayskip}{2pt}%
\setlength{\abovedisplayshortskip}{2pt}%
\setlength{\belowdisplayshortskip}{2pt}%
\setlength{\jot}{0pt}
\begin{align}
w^* = \operatorname*{argmin}_w {f(\mathcal{L}_{\mathcal{M}_u}(w), \mathcal{G}(I(w)))} 
\end{align}
\setlength{\abovedisplayskip}{4pt}%
\setlength{\belowdisplayskip}{4pt}%
\setlength{\abovedisplayshortskip}{4pt}%
\setlength{\belowdisplayshortskip}{4pt}%
\setlength{\jot}{0pt}
in which $\mathcal{L}_{\mathcal{M}_u(w)}$ represents the loss function of the unlearned model, while $\mathcal{G}(\cdot)$ acts as a ``privacy approximation function". This function quantifies how much privacy-sensitive properties of the training data are exposed by the attack features $I(w)$ (e.g., attacking successful rate). The function $f(\cdot)$ represents the linear relationship of the loss function and quantification of privacy leakage, detailed in Section \ref{sec:unlearning}. Therefore, the objective function addresses both the accuracy of the target model and the potential information leakage. To optimize the model weight $w$, we aim to minimize this objective function. This minimization is to ensure the accuracy of the model by reducing $\mathcal{L}_D(w)$ and diminish the leakage of privacy-sensitive data through the attack features $I(w)$.






%

\subsection{Design Goals}
While the retraining approach is an ideal solution in terms of removing the influence of $D_f$ from $\mathcal{M}$, it is not a cost-effective solution. As such, in designing our unlearning mechanism, we consider satisfying the following goals:

\begin{itemize}
\item {\it High Classification Fidelity:} In general, it is expected that any unlearning strategy to adversely impact the accuracy of the target task. This is in part due to the removal of a subset of training samples, potentially including crucial data samples that represent a particular class or category, as well as the removal of feature(s) or even class(es). While such performance degradation is expected, the reduction in accuracy between the unlearned model and the retrained model (or the model before unlearning) should remain minimal.

\item {\it High Unlearning Efficacy:} Unlike retraining and a few exact unlearning approaches, which provide provable guarantees on the unlearned model, other unlearning mechanisms provide bounded approximations at best~\cite{c3}. Nonetheless, any unlearning mechanism should minimize the information that the unlearned model retains from the forget samples -- maintaining high unlearning efficacy. Assessing unlearned model efficacy, beyond theoretical analysis, can be measured via approaches, such as membership inference or uncertainty quantification~\cite{BecLie22}.
\item {\it Data and Model Independent:} The prototypical data, the samples that are representative of classes or features, are more influential in the training of a model. As such, unlearning such samples (even features or classes) would lead to a bigger change in the distribution of the unlearned model. Nonetheless, unlearning strategies should be pertinent to datasets with diverse distributions. A similar property should be held for a diverse set of neural network architectures.
\item {\it Low Latency:} All unlearning strategies require the model's parameters to be updated. Nonetheless, the expected time for transforming $\mathcal{M}$ to $\mathcal{M}_u$, dubbed as unlearning latency, should remain significantly lower than the retraining strategy. The unlearning latency may increase proportional to the fraction of the samples, number of features, and classes that are being removed. 

\end{itemize}
\section{Unlearning Methodology}
\label{sec:methodology}
\begin{figure}[t]
    \centering
   {\includegraphics[width=\columnwidth]{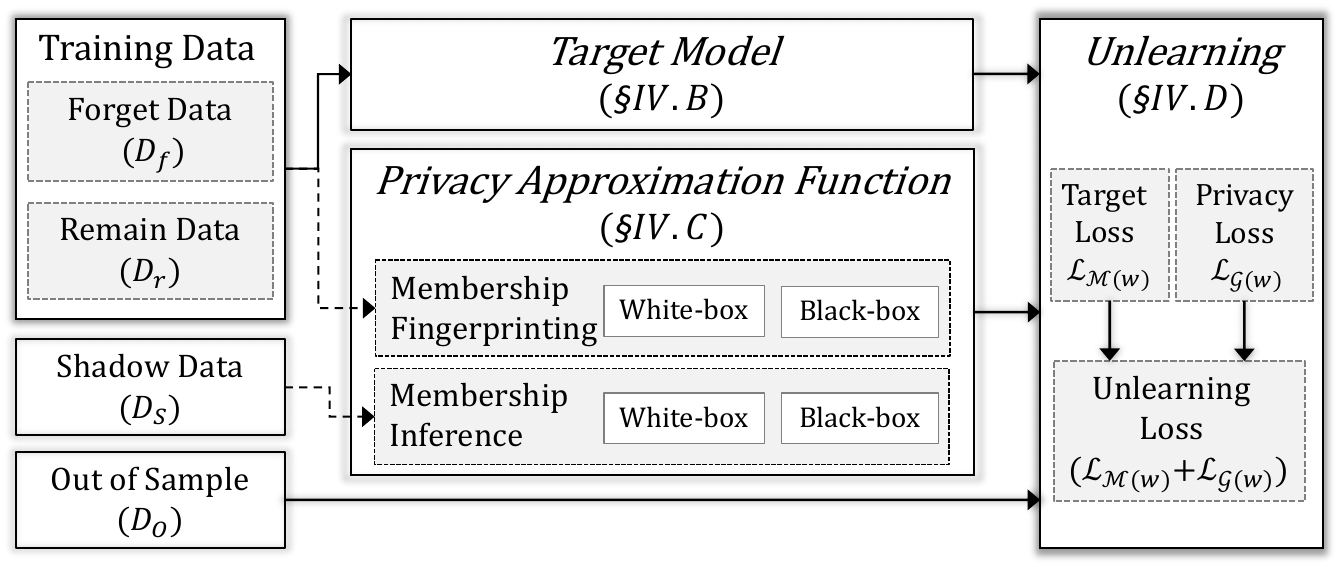}}
    \caption{In \sol, the target model training follows the conventional training approach (\S~\ref{sec:target}). \sol uses a privacy approximation function (\S~\ref{sec:attack}) to infer the privacy-sensitive information of the training data and uses it to guide the unlearning process 
    (\S~\ref{sec:unlearning}). }
    \label{fig:overview}
        \vspace{-5mm}
\end{figure}

In this section, we introduce our machine unlearning mechanism (Figure~\ref{fig:overview}). Our design objective for \sol is to remove information and features associated with $D_f$ through an attack-oriented model optimization. \sol comprises several crucial components, including the configuration of the target model, feature extraction, attack model preparation, and the unlearning optimization process. This unlearning pipeline aims to generate a new, secure model that achieves both high classification fidelity and unlearning efficacy. In what follows, we first provide an overview of our design and then elaborate on the details of each component.

\subsection{Design Overview}
Upon a data removal request, \sol initiates the unlearning process by directly accessing the pre-trained target model $\mathcal{M}$ and the specified data samples $D_f$ that should be forgotten. This allows \sol to complete the unlearning process {\it (i)} independent of the remaining of the original training data, \ie $D_r$, {\it (ii)} independent of the parameters of $\mathcal{M}$ during its training process, or {\it (iii)} without retraining $\mathcal{M}$ from the scratch. Consequently, it effectively addresses the major concerns surrounding machine unlearning.

We consider the classification task for the target model, which follows a given deep learning architecture. We assume the target model $\mathcal{M}$ is trained on an image dataset using a training paradigm, such as stochastic gradient descent (Section~\ref{sec:target}). 
Recall from Section~\ref{sec:privacy} that a privacy approximation function $\mathcal{G}(.)$ exists, which aims to quantify private properties of training data. In \sol, we use function $\mathcal{G}(.)$ in guiding the unlearning process. Function $\mathcal{G}(.)$ can be realized as a neural network and can be trained in various forms for privacy inference. Plausible choices for $\mathcal{G}(.),$ include data reconstruction, property inference, and membership inference functions, which are common techniques in privacy attacks against deep neural networks. In this study, we utilize a membership inference model as the chosen privacy approximation function ($\mathcal{G}(.)$) in the unlearning pipeline. Using the membership inference function, \sol assesses the information leakage of the data samples that clients request to be forgotten during the unlearning process and performs iterative weight refinement until the unlearned model $\mathcal{M}_u$ does not retain any information from $D_f$ (Section~\ref{sec:attack}). 

Upon receiving the removal request for removing $D_f$, we build a customized differentiable unlearning loss function that quantifies the privacy leakage by considering both the information from the target model ($\mathcal{M}$) and the output of the membership inference model, $\mathcal{G}(.)$ (Section~\ref{sec:unlearning}). We then utilize the gradient descent optimization algorithm to update model parameters, aiming to minimize the attack probability distribution between $D_f$ and a non-private \emph{out-of-sample} data, $D_o$, while maintaining classification accuracy without significant reduction. The out-of-sample data is a set of samples with the identical distribution to the training data but with no intersection, \ie $D \cap D_o = \emptyset$. Availability of an out-of-sample distribution is a common assumption in development of responsible AI systems and has been widely used for robust model training~\cite{SriKum18} and inference attacks~\cite{liu2022ml}.

%

To evaluate the effectiveness of our approach, we compare the model generated by our unlearning algorithm ($\mathcal{M}_u$) with benchmark models: trained from scratch and other approximate unlearning algorithms. Furthermore, we illustrate the efficacy of our unlearning approach using a range of evaluation metrics. Lastly, to illustrate the usability of our mechanism, we assess it across several benchmark image datasets, including FMNIST, UTKFace, STL10 and CIFAR-10, on a variety of deep learning architectures, including CNN, ResNet18, Xception, and VGG19.


\subsection{Target Model Training}
\label{sec:target}
This section elaborates on the target model training process. We note that the training of the target model follows the conventional training of deep neural networks and further emphasize that our unlearning process is completely independent of the target model training.
For the training of the machine learning target model, we start with a target dataset $D = \{(x_i, y_i) \mid i \in \{1, \cdots, N\}, y_i \in \{1, \cdots, K\}\}$, where each data point is associated with a class label, $y_i$, belonging to one of the $K$ distinct classes. The primary goal of training the machine learning model $\mathcal{M}$ is to attain high classification performance on dataset $D$, which involves distinguishing among the $K$ classes. The model $\mathcal{M}$ is trained using the cross-entropy loss function, denoted as $\mathcal{L}(w)$, where $w$ represents the model's weights. The loss function measures the dissimilarity between the model's predictions and the actual labels within the dataset $D$, as defined by:
\begin{multline}
  \mathcal{L}_{D}(w) = -\dfrac{1}{N}\sum^{N}_{i=1}\sum^{K}_{j=1}[[y_{i}==j]]\text{log}p\big(y_{i}=j|x_{i};w\big)
\end{multline}
The optimal weights are determined by minimizing the loss function $\mathcal{L}_{D}(w)$ over all training data points $D$, denoted as 
$w^* = \operatorname*{argmin}_w{\mathcal{L}_{D}(w)}$. The weights $w$ is updated through gradient decent steps by $w^t = w^{t-1} - \eta\frac{\partial \mathcal{L}}{\partial w}$. 

During the training process, it is common practice to incorporate out-of-sample data, $D_o$, to facilitate early stopping and prevent overfitting. Instead of adjusting the model weights during target model training, an additional cross-entropy loss, designated as $\mathcal{L}_{D_o}(w)$, is introduced for the purpose of model evaluation.




%
%
\subsection{Privacy Approximation Function}
\label{sec:attack}
In this section, we elaborate on the privacy approximation function, $\mathcal{G(.)}$, which we use in \sol, and the role it plays in the unlearning process. The privacy approximation function, in a general sense, comprises a set of machine learning models, which are utilized to infer the private properties related to the training data through the pre-trained target model $\mathcal{M}$. 
To illustrate this concept, we use a membership inference attack model as an example of $\mathcal{G(.)}$ in designing \sol. Membership inference attacks are commonly employed in assessing privacy vulnerabilities of unlearning models \cite{ChuTarMan23}. The primary objective of $\mathcal{G(.)}$, particularly the MIA model, is to determine whether a particular data sample was part of the training set for $\mathcal{M}$ or not. To make the membership determination, $\mathcal{G(.)}$ relies on a set of features, denoted as $I(w)$, as discussed in Section~\ref{sec:privacy}. These features are derived from the target model $\mathcal{M}$. 
To formalize the concept of MIA\cite{c30}, we consider a set of sensitive data samples denoted as $D_f$ and the attack features $I(w)$ derived from model $\mathcal{M}$. The MIA model, which we henceforth refer to as $\mathcal{G(.)}$, is defined as a binary classifier to produce a class prediction output $\hat{z} = \mathcal{G}(I(w)|_{D_f}) \in \{0,1\}$. A prediction of $\hat{z} = 0$ implies that the data point is not a member of $\mathcal{M}$’s training dataset, while $\hat{z} = 1$ indicates that it is indeed a member of the training dataset.


We consider two distinct types of MIA models: the Black-box attack model and the White-box attack model. In the Black-box attack model, the configuration and parameters of the target model $\mathcal{M}$ are not known to the adversary. Therefore, the primary source of information for deriving attack features used to infer the membership status of the input data is the output of the target model. These attack features include $\hat{p}$, $\hat{y}$, and $\mathcal{L}(\hat{y},y)$ as defined in Section~\ref{sec:privacy}.
In contrast, in the White-box attack, the adversary has access to both, the architecture and the complete set of parameters of the target model $\mathcal{M}$. This accessibility allows the utilization of additional white-box features, including $\nabla \mathcal{L}$ and $[a]_{\mathcal{M}}$. In our approach, we consider both the black-box and white-box attack models as two distinct privacy approximation functions to guide the unlearning process in \sol.


\textbf{Privacy Approximation Function Training}: We train membership inference models, $\mathcal{G}(.)$, using two distinct feature datasets: one for the MIA model and another for the {\it Membership Fingerprinting} (MF) model. The MF model shares similarities with the MIA model in that its purpose is to distinguish between samples that were part of the training set for $\mathcal{M}$ and those that were not. However, it differs from MIA attacks in a fundamental way -- the MF model is trained using the original training data, whereas MIA attacks are typically trained on the shadow dataset since they do not have access to the actual target dataset. 
Among these two approaches, MIA model is a suitable privacy approximation function when access to the training data is limited. Otherwise, the MF model would be a better choice for guiding the unlearning process. 
For both of these models, we offer white-box and black-box variations.
In our unlearning pipeline, adhering to the conventional approach for constructing MIA models, this process involves training a set of shadow models, denoted as $\mathcal{M}_{\text{Shadow}}$, designed to replicate the behavior and characteristics of the target model, $\mathcal{M}$. To train $\mathcal{M}_{\text{Shadow}}$, we utilized the shadow dataset, which is assumed to follow the same underlying distribution as the target dataset but is entirely disjoint. It is worth mentioning that there are diverse methods for constructing shadow datasets~\cite{ShoStrSon17}, and it is commonly assumed that the adversary has access to the shadow dataset~\cite{ShoStrSon17,NasShoHou19,PasAteBer21}.
We further used the architectural design and hyperparameters that we have used for model $\mathcal{M}$ in training $M_{\text{Shadow}}$. These shadow models provide valuable insights into the distribution of the parameter space within the target model, facilitating the inference of private information associated with the data used to train the target model.

To generate attack features for training the MIA-based privacy approximation function, $\mathcal{G}(.)_{\text{MIA}}$, we feed the shadow data into $\mathcal{M}_{\text{Shadow}}$. In the case of the MF-based privacy approximation function, we instead send the original target data to the target model $\mathcal{M}$ to generate the features for training model $\mathcal{G}(.)_{\text{MF}}$. Given that the unlearning process will be conducted by the same party that trained the model, in our evaluation, {\bf we use the white-box MF model as the privacy approximation function to guide the unlearning process.}


\subsection{Neural Network Unlearning}
\label{sec:unlearning}
\begin{figure}[t]
    \centering
    {\includegraphics[width=\columnwidth]{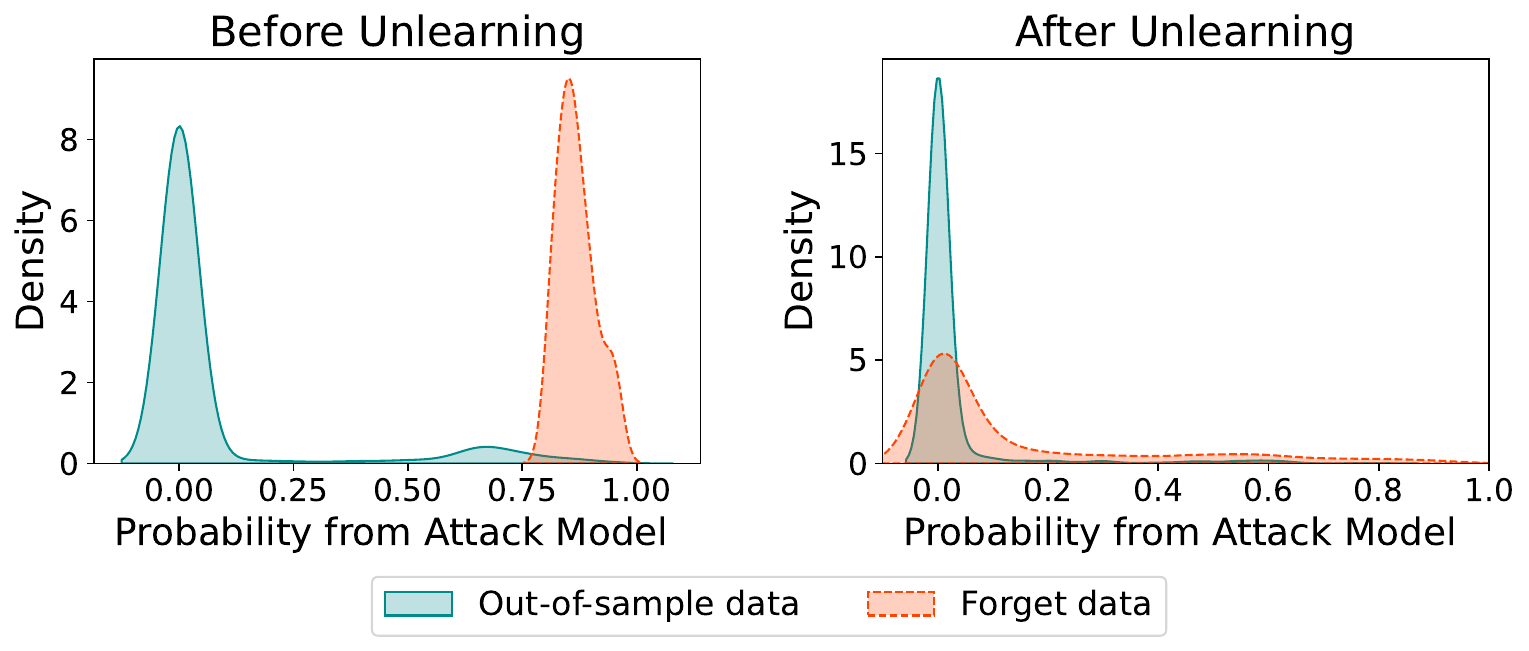}}
    \vspace{-7mm}
    \caption{Membership inference attack probability distributions before and after unlearning. Before unlearning, the MIA attack had a high likelihood of forgetting data and an extremely low likelihood of the out-of-sample data. The objective of our unlearning process is to minimize the divergence between these probability distributions.}
    \label{fig:MIA}
    \vspace{-5mm}
\end{figure}

Given a target training dataset $D = D_f \cup D_r$ and the target model $\mathcal{M}$ (henceforth referred to as $w$) trained on this dataset, we consider that the distributions of attack probabilities generated by the privacy approximation function for forgetting dataset, $\mathcal{G}(I(w)|_{D_f})$, is very high and distinguishable from the attack probabilities for out-of-sample dataset (\eg test dataset). It is worth noting that the choice of out-of-sample data in our work is all non-membership data. Figure~\ref{fig:MIA} illustrates the attack probability distributions on both out-of-sample $D_o$ and forgetting data $D_f$, before and after unlearning. Evidently, the attack shows a low probability for out-of-sample data (\eg the mean value of 0.05\%). In contrast, the attack probability for the forgetting data before unlearning exhibits an exceedingly high likelihood. The primary objective of the unlearning procedure is then to enable the target model $\mathcal{M}$ to forget information $D_f$, resulting in similar attack probability distributions for $D_f$ and $D_o$. Specifically, it aims to ensure that the attackers cannot reconstruct information about $D_f$ using $\mathcal{G}(I(w))$. To formalize the objective goal of unlearning, we have the definition based on the forgetting data and out-of-sample data as follows:

\begin{definition}
   Given the membership inference model $\mathcal{G}(.)$, an optimal unlearning function $S(.)$ for the target model is to make 
   \begin{equation}
    \text{KL}(P(\mathcal{G}(I(S(w))|_{D_f}))||P(\mathcal{G}(I(w)|_{D_o}))) = 0
   \end{equation}
in which $S(w)$ is the model weight updated by unlearning process and $\text{KL}$ is the Kullback-Leibler divergence, $P(\mathcal{G}(I(S(w))|_{D_f}))$ is the attack probabilities on the forgetting data using the unlearned model $S(w)$ and $P(\mathcal{G}(I(w)|_{D_o}))$ is the attack probabilities on the out-of-sample data using the original target model in which the mean of distribution $P(\mathcal{G}(I(w)|_{D_o}))$ is very low.
\label{def:indistinguishable}
\end{definition}

With Definition~\ref{def:indistinguishable}, the goal of a successful unlearning algorithm, $S(.)$, is to update model weights of the target model in such a way that the distance between the distributions of the attack probabilities of the forgetting dataset $D_f$ and the out-of-sample dataset $D_o$ is minimized (\emph{after unlearning} plot in Figure~\ref{fig:MIA}). 
To achieve this, the ideal unlearning algorithm should have the KL divergence between the attack probability distributions of $D_f$ and $D_o$ reduced to zero. Thus, the information of $D_f$ in the dataset can be indistinguishable from the out-of-sample data. 

In addition to unlearning the samples in $D_f$ by minimizing this KL divergence, it is crucial for the target model to maintain accuracy and fidelity in the primary task. In our case, we primarily consider the classification accuracy and cross-entropy loss function for the target model. Therefore, our unlearning approach aims to leverage the combination of classification accuracy (Eqn.~\ref{eqn:accloss}), as measured by cross-entropy loss, and unlearning efficacy (Eqn.~\ref{eqn:priloss}), determined by the KL divergence between attack probability distributions for in-sample and out-of-sample data, during the model update process. 
%
%
We define the loss function to guarantee the classification accuracy of dataset $D_f$ and $D_o$ which is far less than $D$ as:
\begin{equation}
     \mathcal{L}_{\mathcal{M}_u}(w)=\mathcal{L}_{D_f}(w) + \mathcal{L}_{D_o}(w)
     \label{eqn:accloss}
 \end{equation}
and the loss function to guarantee the forgetting data privacy as:
\begin{equation}
    \mathcal{L}_{\mathcal{G}}(w) = \text{KL}(P(\mathcal{G}(I(S(w))|_{D_f}))||P(\mathcal{G}(I(w)|_{D_o})))
    \label{eqn:priloss}
\end{equation}

During the unlearning process, we control the trade-off between the classification accuracy and membership inference attack on $D_f$ by assigning the multipliers $\lambda_1$ to $\mathcal{L}_{\mathcal{M}}$ and $\lambda_2$ to $\mathcal{L}_{\mathcal{G}}$. These multipliers are used to balance the impact of the two loss functions on the overall learning process. The final loss function to get the optimal weight $w^*$ for the unlearned model can be formalized as:
\begin{equation}
w^* = \operatorname*{argmin}_w{\{\lambda_{1}\times\mathcal{L}_{\mathcal{M}_u}(w) + \lambda_{2}\times\mathcal{L}_{\mathcal{G}} (w)}\}
\end{equation}

To simplify the loss function to get the optimal model weights, we can consider the upper bound of the $\mathcal{L}_{\mathcal{G}}(w)$. By definition of the KL divergence, and considering that the probability of attack should be less than 1, we can hold the following inequality:
\begin{align}
     \mathcal{L}_{\mathcal{G}}(w) &= \text{KL}(P(\mathcal{G}(I(S(w))|_{D_f})||P(\mathcal{G}(I(w)|_{D_o})) \nonumber \\
     &\leq \log(\frac{P(\mathcal{G}(I(S(w))|_{D_f}))}{P(\mathcal{G}(I(w)|_{D_o}))})
\end{align}
Based on the distribution observation in Figure \ref{fig:MIA}, we assume that the distributions of attack based on the dataset $D_f$ and $D_o$ satisfy the Gaussian distribution as follows:

\begin{equation*}
   \mathcal{G}(I(S(w))|_{D_f}) \sim N(\mu_{D_f}, {\sigma}^2) 
\end{equation*}
\begin{equation*}
   \mathcal{G}(I(w)|_{D_o}) \sim N(\mu_{D_o}, {\sigma}^2) 
\end{equation*}
\begin{equation}
   P(\mathcal{G}(I(S(w))|_{D_f})) = \frac{1}{\sigma\sqrt{2\pi}}exp^{-\frac{1}{2}(\frac{\mathcal{G}(I(S(w))|_{D_f})-\mu_{D_f}}{\sigma})^2} 
\end{equation}
\begin{equation}
  P(\mathcal{G}(I(w)|_{D_o})) = \frac{1}{\sigma\sqrt{2\pi}}exp^{-\frac{1}{2}(\frac{\mathcal{G}(I(w)|_{D_o})-\mu_{D_o}}{\sigma})^2}  
\end{equation}
%
%
Thus, the upper bound of $\mathcal{L}_{\mathcal{G}}(w)$ can be expressed as:


\begin{align}
     \mathcal{L}_{\mathcal{G}}(w) 
     &\leq \log(\frac{P(\mathcal{G}(I(S(w))|_{D_f}))}{P(\mathcal{G}(I(w)|_{D_o}))}) \nonumber \\ 
  &\leq -\frac{1}{2}(\frac{\mathcal{G}(I(S(w))|_{D_f})-\mu_{D_f}}{\sigma})^2 \nonumber \\ &+ \frac{1}{2}(\frac{\mathcal{G}(I(w)|_{D_o}) - \mu_{D_o}}{\sigma})^2
\end{align}

Recall that $\mathcal{G}(I(w)|_{D_o})$ is fixed given that the target model is known and remains unchanged during the unlearning process. Thus, when we minimize the value of $\mathcal{L}_{\mathcal{G}}(w)$ to guarantee the robustness against the membership inference attack on forgetting data $D_f$, the objective function is equivalent to minimizing the $\mathcal{L}_{\mathcal{G}}(w) = -\log\big(\frac{\mathcal{G}(I(S(w))|_{D_f})-\mu_{D_f}}{\sigma}\big)^2$. We take the logarithm in order to enable the loss function $\mathcal{L}_{\mathcal{G}}(w)$ to grow larger when $|\mathcal{G}(I(S(w))|_{D_f})-\mu_{D_f}| \rightarrow 0$.  
Moreover, since the goal of the unlearning process is to decrease the attack probability of forgetting data, we would have one condition that
\begin{equation}
    \mathcal{G}(I(S(w))|_{D_f}) \leq \mu_{D_f} \leq 1 \nonumber
\end{equation}
Thus, our problem is equivalent to minimizing:
\begin{equation}
-\log(\mu_{D_f}-\mathcal{G}(I(S(w))|_{D_f})) \nonumber
\label{eqn:11}
\end{equation}
As the probability of attack for the forgetting dataset $D_f$ is close to $1$, minimizing the previous term is equivalent to minimizing:
\begin{equation}
-\log(1-\mathcal{G}(I(S(w))|_{D_f})) 
\end{equation}

Hence, in \sol, we define the loss function for unlearning the target model as: 
%
\begin{align}
  \mathcal{L} 
    &=\lambda_1 \times \underbrace{(\mathcal{L}_{D_f}(w) + \mathcal{L}_{D_o}(w))}_{\mathcal{L}_{\mathcal{M}_u}(w)} \nonumber \\ &+ 
     \lambda_2 \times \underbrace{(-\log (1-\mathcal{G}(I(S(w))|_{D_f})))}_{\mathcal{L}_\mathcal{G}(w)}  
\end{align}

The optimal weights of the unlearned model are defined as:
\begin{align}
w^* = \operatorname*{argmin}_w & \{\ \lambda_1 \times {\mathcal{L}_{\mathcal{M}_u}(w)} + \lambda_2 \times {\mathcal{L}_\mathcal{G}(w)} \}
\label{eqn:losstol}
\end{align}

Minimizing this first term ($\mathcal{L}_{\mathcal{M}_u}(w)$) is to minimize the loss function retaining the accuracy of the unlearned model for forgetting and out-of-sample data. Minimizing the second term ($\mathcal{L}_\mathcal{G}(w)$) is to decrease the amount of sensitive information extracted on $D_f$ by the privacy approximation function $\mathcal{G}(.)$ from the unlearned model. 
The combination of $\mathcal{L}_{\mathcal{M}_u}(w)$ and $\mathcal{L}_\mathcal{G}(w)$ is employed to update the weights of the target model to derive the unlearned model, $S(w)$. By updating the weights, the attack accuracy on forgetting data $D_f$ should reduce as a result of minimizing $\mathcal{L}_\mathcal{G}(w)$ while still preserving the classification accuracy by minimizing  $\mathcal{L}_{\mathcal{M}_u}(w)$, which we will validate in our evaluation (detailed in Table~\ref{tab:unlearn_target_acc} and Table~\ref{tab:unlearn_MIA_acc}).
Next, we will assess the performance of the unlearned model using metrics, including classification fidelity, unlearning efficacy, and unlearning latency.

\section{EXPERIMENTS AND RESULTS}


\subsection{Experimental Workflows}

We used the following datasets to evaluate \sol:
\begin{itemize}
    \item {FMNIST} dataset~\cite{xiao2017fashion} consisting a comprehensive set of 70,000 grayscale images. These images encompass a diverse array of 10 fashion items, including various types of clothing, dresses, shoes, handbags, and more. 
    %
    \item {UTKFace}~\cite{c36} dataset comprises over 23,000 face images annotated with age, gender, and ethnicity, displaying 
    varied pose, facial expression, illumination, occlusion, and resolution. 
    \item {STL10}~\cite{coates2011analysis} dataset comprises 13,000 images in 10 classes: airplane, bird, car, cat, deer, dog, horse, monkey, ship, and truck.
    \item {CIFAR-10}~\cite{CIFAR} dataset features 60K color images ($32\times32$ pixels) across 10 classes, with 6K images per class. There are $50000$ training and $10000$ test images.
\end{itemize}
We test the efficacy of \sol on four different architectures: SimpleCNN, ResNet18 \cite{he2016deep}, Xception \cite{chollet2017xception}, and VGG19 \cite{simonyan2014very}.
The SimpleCNN model (hereafter CNN) consists of 3 convolutional layers, each followed by a ReLU activation function and a max-pooling layer, along with two fully connected layers for classification. 
For the Xception architecture, we have implemented the same model as presented in the original paper~\cite{chollet2017xception}. We have adopted the ResNet model with 18 layers, as detailed in~\cite{he2016deep}. The VGG19 model we used consists of a total of 19 layers, including 16 convolutional layers, 3 fully connected layers, and ReLU as its activation function~\cite{simonyan2014very}. All four deep learning architectures employ a softmax classifier in the output layer to predict probabilities associated with distinct classes. The network configurations and training procedures are implemented in PyTorch. Each target model undergoes training for 50 epochs using the SGD optimizer with a learning rate of 1e-2, a momentum of 0.9, and a weight decay of 5e-4.

We divided each dataset into four equal-sized subsets. We use two subsets to train our target model, and the remaining two subsets are treated as shadow distribution, which will be utilized to conduct a membership inference attack on the target model. All the image samples in the datasets are pre-processed before using them for model training, including re-sizing (64$\times$64) and normalization.

%

In \sol we use the membership fingerprinting model to conduct unlearning. Then, we assess its effectiveness through a membership inference attack conducted before and after unlearning in white-box and black-box settings. The MIA utilizes the framework proposed in prior work and assumes the same threat model~\cite{liu2022ml}.

Subsequently, we test \sol on four \emph{forgetting datasets}, each with an increasing number of target samples selected evenly from each class that were classified as training data by the MF model with high probability. The sizes of these forgetting datasets are 0.01, 0.1, 0.25, and 0.5 of the target training datasets, respectively.
Table~\ref{tab:dataset_statistics} summarizes the number of data points in the forgetting data curated from each of the four datasets. 

Given a pre-trained target model, we apply our unlearning mechanism (\sol) to update the parameters of the target model 
for each forgetting dataset. In particular, \sol refines the parameters of the target model by minimizing the classification loss and unlearning loss based on the forgetting dataset, as elaborated in Section~\ref{sec:unlearning}. We also benchmark our algorithm with two unlearning approaches: Naive Retraining \cite{thudi2022unrolling} and Fisher Unlearning \cite{BecLie22}. 
The Naive Retraining algorithm retrains the target model on the remaining data by excluding specific forgetting data from the training dataset, while maintaining consistent hyper-parameter configurations similar to those used in the original training process. In the case of Fisher Unlearning, we followed the same implementation outlined in prior research to ensure the consistency of unlearning performance \cite{BecLie22}.



\begin{table}[tbp]
  \centering
  \caption{Statistics of data in experiment dataset.}
        \vspace{-3mm}
    \begin{tabular}{|c|c|c|c|c|c|c|}
    \hline
          & \multicolumn{2}{c|}{Target Dataset} & \multicolumn{4}{c|}{Forgetting dataset} \\
    \hline
          & Train & Test  & 0.01  & 0.1   & 0.25  & 0.5 \\
    \hline
    FMNIST &17500       &17500       &179   & 1755      &4380  &8753  \\
    \hline
    UTKFace &5503       &5503       &57       &552       &1378   &2753  \\
    \hline
    STL10 &3250       &3250       &40       &331       &818       &1627  \\
    \hline
    CIFAR-10 &15000       &15000       &154       &1504       &3754       &7504  \\
    \hline
    \end{tabular}%
    \label{tab:dataset_statistics}%
    \vspace{-6mm}
\end{table}%

To assess the robustness of our unlearning framework, each target model is trained using 20 different initializations (seed numbers), resulting in a total of 80 trained models. Each of these models is then assessed in both black-box and white-box settings. Following the implementation of white-box MIA models outlined in~\cite{liu2022ml}, we selected four attack features from the target model to serve as inputs to out MF and MIA models. Per Section \ref{sec:privacy}, these features include
$\text{$I(w) \subset \{\hat{p}, \hat{y}, \mathcal{L}(\hat{y},y), \nabla \mathcal{L}\}$}.$ 
The white-box attack model was trained for 50 epochs using the Adam optimizer with a learning rate of 1e-5. For the black-box attack, we used the MIA implementation from IBM's Adversarial Robustness Toolbox~\cite{nicolae2018adversarial}.

Upon completion of the target model and the attack model, we initiate the unlearning process using the specified forgetting data as the input. In our unlearning framework, we explore various hyperparameter values that are used in \sol's loss function, as outlined in Equation~\ref{eqn:losstol}, including the unlearning weight $\lambda_2$ and the learning rate $\eta$ during the unlearning process. It's worth noting that in our experimental setup, we defined $\lambda_1 = 1 - \lambda_2$. To ensure optimal unlearning performance, we identified the most suitable values for these hyperparameters across all deep learning architectures and datasets. The details of these optimal hyperparameter values are available in Table~\ref{tab:hyperparameter_unlearning} in the supplementary document.


\begin{figure}[!tbp] 
    \centering
    \includegraphics[width=\columnwidth]{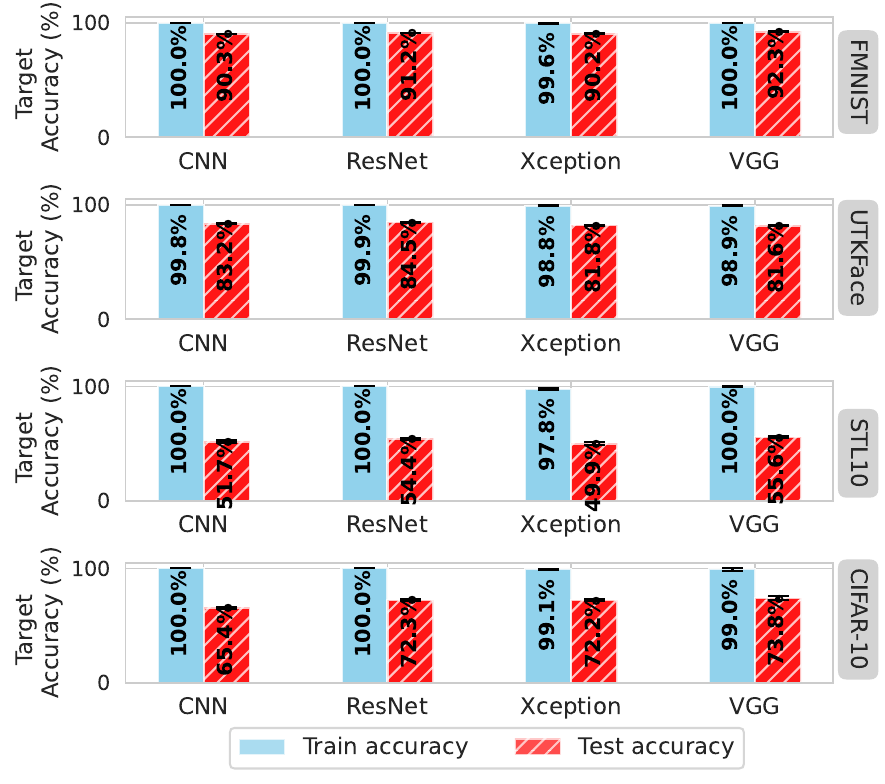}
    \vspace{-7mm}
    \caption{The train and test classification accuracy of four datasets across four deep learning architectures before unlearning.}
    \vspace{-4mm}
    \label{fig:accuracy}
\end{figure}

\subsection{Results}
\label{sec:experiment}
\textbf{Target Model Evaluation}: We first present the training and test accuracy of the target models based on different neural network architectures trained on four datasets. Figure~\ref{fig:accuracy} provides insight into the dataset complexities used in our experiments. Notably, the FMNIST dataset yields the highest overall test classification accuracy across all architectures, with CNN (90.3\%), ResNet18 (91.2\%), Xception (90.2\%), and VGG19 (92.3\%). In the case of the UTKFace and CIFAR-10 datasets, both training and test accuracies achieve lower averaged classification accuracy when compared to the four deep learning architectures trained on the FMNIST dataset. Conversely, the STL10 dataset exhibits the lowest test accuracy among the three datasets for CNN (51.7\%), ResNet18 (54.4\%), Xception (49.9\%), and VGG (55.6\%), showing its greater complexity in terms of classification challenges.

\begin{figure}[!tbp]
    \centering
    {\includegraphics[width=\columnwidth]{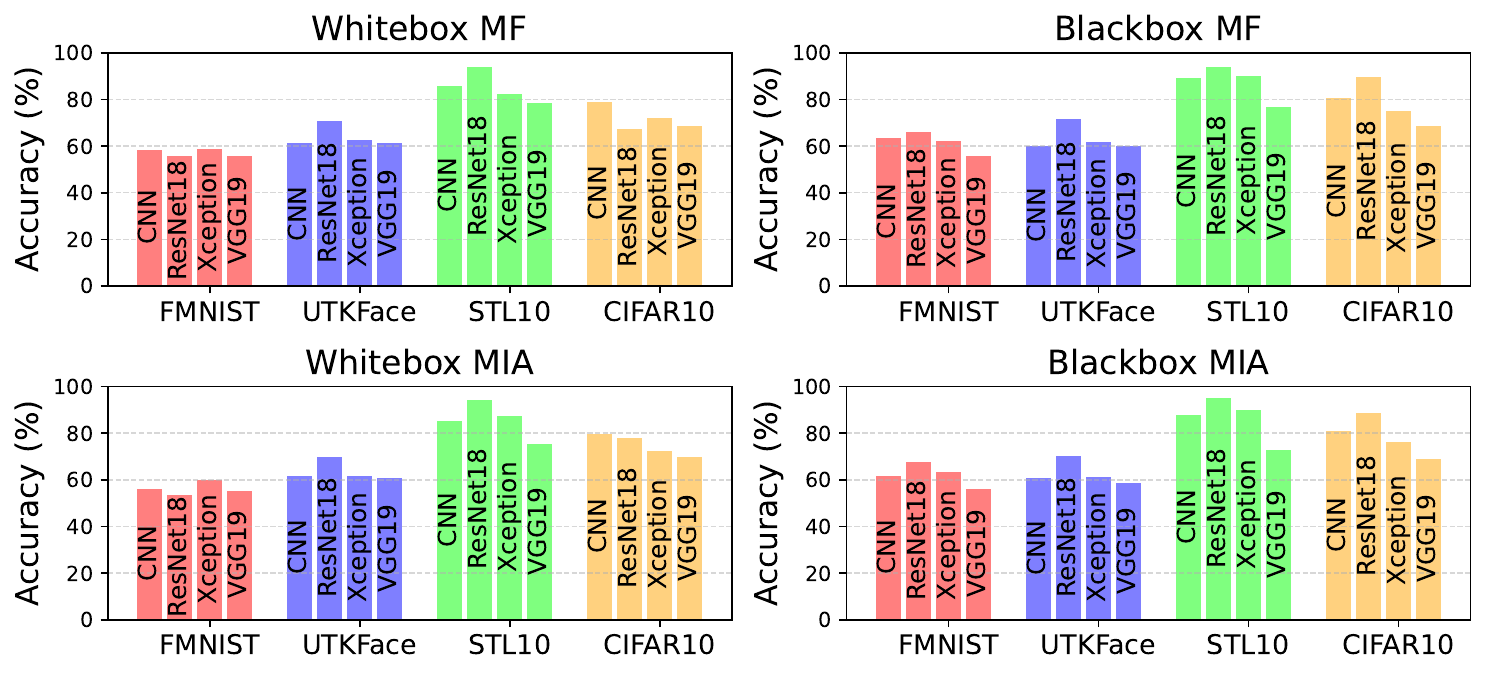}}
    \vspace{-7mm}
    \caption{Accuracy of the privacy approximation functions (MIA and MF) before unlearning in both white-box and black-box forms across all datasets. The results suggest that MIA and MF models perform similarly. Thus, either of these models can be used as the privacy approximation function to guide \sol's unlearning process.}
    \label{fig:4b}
    \vspace{-2mm}
\end{figure}

\noindent
\textbf{Privacy Approximation Function Evaluation}: We further evaluate the performance of two privacy approximation functions, \ie MIA and MF, on inferring whether a data sample belongs to the training datasets or not in both black-box and white-box settings. The outcome of this experiment shed light on the suitability of MIA and MF for guiding \sol unlearning process. Figure~\ref{fig:4b} demonstrates that MIA and MF yield similar results in both black-box and white-box settings. Notably, the results indicate that MIA and MF achieve higher prediction accuracy on the STL10 dataset, which is partly due to the train and test accuracy gap in the target classification task (Figure~\ref{fig:accuracy}). These observations align with previous research findings~\cite{liu2022ml}, suggesting that membership inference (fingerprinting) accuracy tends to be higher when the target models experience overfitting. More importantly, these results suggest that utilizing MIA or MF will lead to a similar unlearning performance. Hence, given the performance similarities of these models and considering that training the MF model is less cumbersome for the model owner, we use MF to guide \sol unlearning process.

\noindent
\textbf{Unlearning Evaluation}: 
In this section, we will evaluate \sol in terms of classification fidelity and unlearning efficacy.
To assess the classification fidelity of the unlearning process, we evaluate the difference in the classification accuracy of the target model before and after unlearning.
Table~\ref{tab:unlearn_target_acc} presents a comprehensive summary of the predictive accuracy of the training (\ie forgetting data samples ($D_f$) and the remaining training data ($D_r$)) and the test phases, both before and after unlearning. 
%
The results indicate that the unlearned model consistently maintains accuracy levels comparable to the original target model across all architectures and datasets. After the unlearning, we observe minimal changes in classification performance compared to the target model, with a maximum accuracy drop of less than 10\% (CIFAR10 on VGG19). Interestingly, one can observe that in some cases (STL10 on ResNet18, Xception, and VGG19), \sol has marginally improved the test accuracy of the unlearned models. This accuracy boost is due to the higher generalizability of the unlearned model as a result of using out-of-sample data during the unlearning process.

\begin{table}[tbp]
  \centering
  \caption{Classification accuracy of target models before and after unlearning test data.}
  \vspace{-0mm}
    \begin{tabular}{|c|c|l|c|c|c|c|}
    \hline
     \multicolumn{1}{|l|}{} & \multicolumn{1}{l|}{\textbf{Df}} & \textbf{Phase} & \multicolumn{1}{l|}{\textbf{UTKFace}} & \multicolumn{1}{l|}{\textbf{FMNIST}} & \multicolumn{1}{l|}{\textbf{STL10}} & \multicolumn{1}{l|}{\textbf{CIFAR10}} \\
    \hline
    \multirow{8}[0]{*}{\begin{sideways}\rotatebox[origin=c]{270}{CNN}\end{sideways}} & \multirow{2}[1]{*}{1\%} & Before & 0.83  & 0.9   & 0.48  & 0.65 \\
\cline{3-7}          &       & \cellcolor[rgb]{ .816,  .816,  .816}After & \cellcolor[rgb]{ .816,  .816,  .816}0.82 & \cellcolor[rgb]{ .816,  .816,  .816}0.89 & \cellcolor[rgb]{ .816,  .816,  .816}0.49 & \cellcolor[rgb]{ .816,  .816,  .816}0.61 \\
\cline{2-7}          & \multirow{2}[1]{*}{10\%} & Before & 0.82  & 0.9   & 0.49  & 0.66 \\
\cline{3-7}          &       & \cellcolor[rgb]{ .816,  .816,  .816}After & \cellcolor[rgb]{ .816,  .816,  .816}0.8 & \cellcolor[rgb]{ .816,  .816,  .816}0.89 & \cellcolor[rgb]{ .816,  .816,  .816}0.52 & \cellcolor[rgb]{ .816,  .816,  .816}0.63 \\
\cline{2-7}          & \multirow{2}[1]{*}{25\%} & Before & 0.83  & 0.91  & 0.51  & 0.66 \\
\cline{3-7}          &       & \cellcolor[rgb]{ .816,  .816,  .816}After & \cellcolor[rgb]{ .816,  .816,  .816}0.79 & \cellcolor[rgb]{ .816,  .816,  .816}0.89 & \cellcolor[rgb]{ .816,  .816,  .816}0.5 & \cellcolor[rgb]{ .816,  .816,  .816}0.56 \\
\cline{2-7}          & \multirow{2}[1]{*}{50\%} & Before & 0.84  & 0.91  & 0.52  & 0.65 \\
\cline{3-7}          &       & \cellcolor[rgb]{ .816,  .816,  .816}After & \cellcolor[rgb]{ .816,  .816,  .816}0.8 & \cellcolor[rgb]{ .816,  .816,  .816}0.89 & \cellcolor[rgb]{ .816,  .816,  .816}0.5 & \cellcolor[rgb]{ .816,  .816,  .816}0.6 \\
    \hline
    \multirow{8}[0]{*}{\begin{sideways}\rotatebox[origin=c]{270}{ResNet18}\end{sideways}} & \multirow{2}[1]{*}{1\%} & Before & 0.84  & 0.91  & 0.55  & 0.72 \\
\cline{3-7}          &       & \cellcolor[rgb]{ .816,  .816,  .816}After & \cellcolor[rgb]{ .816,  .816,  .816}0.82 & \cellcolor[rgb]{ .816,  .816,  .816}0.91 & \cellcolor[rgb]{ .816,  .816,  .816}0.55 & \cellcolor[rgb]{ .816,  .816,  .816}0.72 \\
\cline{2-7}          & \multirow{2}[1]{*}{10\%} & Before & 0.83  & 0.91  & 0.57  & 0.71 \\
\cline{3-7}          &       & \cellcolor[rgb]{ .816,  .816,  .816}After & \cellcolor[rgb]{ .816,  .816,  .816}0.83 & \cellcolor[rgb]{ .816,  .816,  .816}0.91 & \cellcolor[rgb]{ .816,  .816,  .816}0.58 & \cellcolor[rgb]{ .816,  .816,  .816}0.7 \\
\cline{2-7}          & \multirow{2}[1]{*}{25\%} & Before & 0.86  & 0.91  & 0.54  & 0.72 \\
\cline{3-7}          &       & \cellcolor[rgb]{ .816,  .816,  .816}After & \cellcolor[rgb]{ .816,  .816,  .816}0.85 & \cellcolor[rgb]{ .816,  .816,  .816}0.9 & \cellcolor[rgb]{ .816,  .816,  .816}0.57 & \cellcolor[rgb]{ .816,  .816,  .816}0.71 \\
\cline{2-7}          & \multirow{2}[1]{*}{50\%} & Before & 0.85  & 0.91  & 0.59  & 0.73 \\
\cline{3-7}          &       & \cellcolor[rgb]{ .816,  .816,  .816}After & \cellcolor[rgb]{ .816,  .816,  .816}0.85 & \cellcolor[rgb]{ .816,  .816,  .816}0.91 & \cellcolor[rgb]{ .816,  .816,  .816}0.59 & \cellcolor[rgb]{ .816,  .816,  .816}0.72 \\
    \hline
    \multirow{8}[0]{*}{\begin{sideways}\rotatebox[origin=c]{270}{Xception}\end{sideways}} & \multirow{2}[1]{*}{1\%} & Before & 0.82  & 0.92  & 0.54  & 0.76 \\
\cline{3-7}          &       & \cellcolor[rgb]{ .816,  .816,  .816}After & \cellcolor[rgb]{ .816,  .816,  .816}0.81 & \cellcolor[rgb]{ .816,  .816,  .816}0.92 & \cellcolor[rgb]{ .816,  .816,  .816}0.54 & \cellcolor[rgb]{ .816,  .816,  .816}0.77 \\
\cline{2-7}          & \multirow{2}[1]{*}{10\%} & Before & 0.83  & 0.93  & 0.57  & 0.76 \\
\cline{3-7}          &       & \cellcolor[rgb]{ .816,  .816,  .816}After & \cellcolor[rgb]{ .816,  .816,  .816}0.83 & \cellcolor[rgb]{ .816,  .816,  .816}0.93 & \cellcolor[rgb]{ .816,  .816,  .816}0.56 & \cellcolor[rgb]{ .816,  .816,  .816}0.75 \\
\cline{2-7}          & \multirow{2}[1]{*}{25\%} & Before & 0.81  & 0.93  & 0.54  & 0.76 \\
\cline{3-7}          &       & \cellcolor[rgb]{ .816,  .816,  .816}After & \cellcolor[rgb]{ .816,  .816,  .816}0.78 & \cellcolor[rgb]{ .816,  .816,  .816}0.92 & \cellcolor[rgb]{ .816,  .816,  .816}0.53 & \cellcolor[rgb]{ .816,  .816,  .816}0.77 \\
\cline{2-7}          & \multirow{2}[1]{*}{50\%} & Before & 0.81  & 0.92  & 0.55  & 0.76 \\
\cline{3-7}          &       & \cellcolor[rgb]{ .816,  .816,  .816}After & \cellcolor[rgb]{ .816,  .816,  .816}0.76 & \cellcolor[rgb]{ .816,  .816,  .816}0.92 & \cellcolor[rgb]{ .816,  .816,  .816}0.57 & \cellcolor[rgb]{ .816,  .816,  .816}0.77 \\
    \hline
    \multirow{8}[0]{*}{\begin{sideways}\rotatebox[origin=c]{270}{VGG19}\end{sideways}} & \multirow{2}[1]{*}{1\%} & Before & 0.83  & 0.88  & 0.52  & 0.72 \\
\cline{3-7}          &       & \cellcolor[rgb]{ .816,  .816,  .816}After & \cellcolor[rgb]{ .816,  .816,  .816}0.84 & \cellcolor[rgb]{ .816,  .816,  .816}0.9 & \cellcolor[rgb]{ .816,  .816,  .816}0.57 & \cellcolor[rgb]{ .816,  .816,  .816}0.62 \\
\cline{2-7}          & \multirow{2}[1]{*}{10\%} & Before & 0.82  & 0.89  & 0.48  & 0.72 \\
\cline{3-7}          &       & \cellcolor[rgb]{ .816,  .816,  .816}After & \cellcolor[rgb]{ .816,  .816,  .816}0.83 & \cellcolor[rgb]{ .816,  .816,  .816}0.87 & \cellcolor[rgb]{ .816,  .816,  .816}0.49 & \cellcolor[rgb]{ .816,  .816,  .816}0.6 \\
\cline{2-7}          & \multirow{2}[1]{*}{25\%} & Before & 0.83  & 0.9   & 0.47  & 0.73 \\
\cline{3-7}          &       & \cellcolor[rgb]{ .816,  .816,  .816}After & \cellcolor[rgb]{ .816,  .816,  .816}0.83 & \cellcolor[rgb]{ .816,  .816,  .816}0.91 & \cellcolor[rgb]{ .816,  .816,  .816}0.49 & \cellcolor[rgb]{ .816,  .816,  .816}0.61 \\
\cline{2-7}          & \multirow{2}[1]{*}{50\%} & Before & 0.8   & 0.9   & 0.5   & 0.73 \\
\cline{3-7}          &       & \cellcolor[rgb]{ .816,  .816,  .816}After & \cellcolor[rgb]{ .816,  .816,  .816}0.82 & \cellcolor[rgb]{ .816,  .816,  .816}0.9 & \cellcolor[rgb]{ .816,  .816,  .816}0.52 & \cellcolor[rgb]{ .816,  .816,  .816}0.75 \\
    \hline
    \end{tabular}%
  \label{tab:unlearn_target_acc}%
  \vspace{-2mm}
\end{table}%

\begin{table}[!t]
  \centering
  \caption{Whitebox MIA accuracy of unlearned models before and after unlearning on forget data.}
  \vspace{-0mm}
\begin{tabular}{|c|c|l|c|c|c|c|}
\hline
 \multicolumn{1}{|l|}{} & \multicolumn{1}{l|}{\textbf{Df}} & \textbf{Phase} & \multicolumn{1}{l|}{\textbf{UTKFace}} & \multicolumn{1}{l|}{\textbf{FMNIST}} & \multicolumn{1}{l|}{\textbf{STL10}} & \multicolumn{1}{l|}{\textbf{CIFAR10}} \\
\hline
\multirow{8}[0]{*}{\begin{sideways}\rotatebox[origin=c]{270}{CNN}\end{sideways}} 
& \multirow{2}[1]{*}{1\%} & Before & 1.00 & 1.00 & 0.975 & 1.00 \\
\cline{3-7}          &       & \cellcolor[rgb]{ .816,  .816,  .816}After & \cellcolor[rgb]{ .816,  .816,  .816}0.052 & \cellcolor[rgb]{ .816,  .816,  .816}0.318 & \cellcolor[rgb]{ .816,  .816,  .816}0.15 & \cellcolor[rgb]{ .816,  .816,  .816}0.019 \\
\cline{2-7}          & \multirow{2}[1]{*}{10\%} & Before & 0.981 & 1.00 & 0.975 & 1.00 \\
\cline{3-7}          &       & \cellcolor[rgb]{ .816,  .816,  .816}After & \cellcolor[rgb]{ .816,  .816,  .816}0.106 & \cellcolor[rgb]{ .816,  .816,  .816}0.365 & \cellcolor[rgb]{ .816,  .816,  .816}0.166 & \cellcolor[rgb]{ .816,  .816,  .816}0.013 \\
\cline{2-7}          & \multirow{2}[1]{*}{25\%} & Before & 0.962 & 1.00 & 0.979 & 1.00 \\
\cline{3-7}          &       & \cellcolor[rgb]{ .816,  .816,  .816}After & \cellcolor[rgb]{ .816,  .816,  .816}0.129 & \cellcolor[rgb]{ .816,  .816,  .816}0.485 & \cellcolor[rgb]{ .816,  .816,  .816}0.106 & \cellcolor[rgb]{ .816,  .816,  .816}0.026 \\
\cline{2-7}          & \multirow{2}[1]{*}{50\%} & Before & 0.941 & 1.00 & 0.982 & 0.999 \\
\cline{3-7}          &       & \cellcolor[rgb]{ .816,  .816,  .816}After & \cellcolor[rgb]{ .816,  .816,  .816}0.19 & \cellcolor[rgb]{ .816,  .816,  .816}0.588 & \cellcolor[rgb]{ .816,  .816,  .816}0.101 & \cellcolor[rgb]{ .816,  .816,  .816}0.044 \\
\hline
\multirow{8}[0]{*}{\begin{sideways}\rotatebox[origin=c]{270}{ResNet18}\end{sideways}} 
& \multirow{2}[1]{*}{1\%} & Before & 1.00 & 1.00 & 1.00 & 1.00 \\
\cline{3-7}          &       & \cellcolor[rgb]{ .816,  .816,  .816}After & \cellcolor[rgb]{ .816,  .816,  .816}0.052 & \cellcolor[rgb]{ .816,  .816,  .816}0 & \cellcolor[rgb]{ .816,  .816,  .816}0.1 & \cellcolor[rgb]{ .816,  .816,  .816}0.006 \\
\cline{2-7}          & \multirow{2}[1]{*}{10\%} & Before & 1.00 & 0.997 & 1.00 & 1.00 \\
\cline{3-7}          &       & \cellcolor[rgb]{ .816,  .816,  .816}After & \cellcolor[rgb]{ .816,  .816,  .816}0.001 & \cellcolor[rgb]{ .816,  .816,  .816}0.002 & \cellcolor[rgb]{ .816,  .816,  .816}0.009 & \cellcolor[rgb]{ .816,  .816,  .816}0.001 \\
\cline{2-7}          & \multirow{2}[1]{*}{25\%} & Before & 1.00 & 0.997 & 1.00 & 0.998 \\
\cline{3-7}          &       & \cellcolor[rgb]{ .816,  .816,  .816}After & \cellcolor[rgb]{ .816,  .816,  .816}0 & \cellcolor[rgb]{ .816,  .816,  .816}0.002 & \cellcolor[rgb]{ .816,  .816,  .816}0.004 & \cellcolor[rgb]{ .816,  .816,  .816}0.0005 \\
\cline{2-7}          & \multirow{2}[1]{*}{50\%} & Before & 1.00 & 0.987 & 1.00 & 0.982 \\
\cline{3-7}          &       & \cellcolor[rgb]{ .816,  .816,  .816}After & \cellcolor[rgb]{ .816,  .816,  .816}0 & \cellcolor[rgb]{ .816,  .816,  .816}0.001 & \cellcolor[rgb]{ .816,  .816,  .816}0.007 & \cellcolor[rgb]{ .816,  .816,  .816}0.001 \\
\hline
\multirow{8}[0]{*}{\begin{sideways}\rotatebox[origin=c]{270}{Xception}\end{sideways}} 
& \multirow{2}[1]{*}{1\%} & Before & 1.00 & 0.977 & 1.00 & 1.00 \\
\cline{3-7}          &       & \cellcolor[rgb]{ .816,  .816,  .816}After & \cellcolor[rgb]{ .816,  .816,  .816}0.21 & \cellcolor[rgb]{ .816,  .816,  .816}0.005 & \cellcolor[rgb]{ .816,  .816,  .816}0.47 & \cellcolor[rgb]{ .816,  .816,  .816}0 \\
\cline{2-7}          & \multirow{2}[1]{*}{10\%} & Before & 1.00 & 0.879 & 1.00 & 1.00 \\
\cline{3-7}          &       & \cellcolor[rgb]{ .816,  .816,  .816}After & \cellcolor[rgb]{ .816,  .816,  .816}0.21 & \cellcolor[rgb]{ .816,  .816,  .816}0.002 & \cellcolor[rgb]{ .816,  .816,  .816}0.314 & \cellcolor[rgb]{ .816,  .816,  .816}0 \\
\cline{2-7}          & \multirow{2}[1]{*}{25\%} & Before & 0.929 & 0.883 & 1.00 & 0.999 \\
\cline{3-7}          &       & \cellcolor[rgb]{ .816,  .816,  .816}After & \cellcolor[rgb]{ .816,  .816,  .816}0.156 & \cellcolor[rgb]{ .816,  .816,  .816}0.002 & \cellcolor[rgb]{ .816,  .816,  .816}0.067 & \cellcolor[rgb]{ .816,  .816,  .816}0.0002 \\
\cline{2-7}          & \multirow{2}[1]{*}{50\%} & Before & 0.849 & 0.884 & 1.00 & 1.00 \\
\cline{3-7}          &       & \cellcolor[rgb]{ .816,  .816,  .816}After & \cellcolor[rgb]{ .816,  .816,  .816}0.039 & \cellcolor[rgb]{ .816,  .816,  .816}0.004 & \cellcolor[rgb]{ .816,  .816,  .816}0.031 & \cellcolor[rgb]{ .816,  .816,  .816}0 \\
\hline
\multirow{8}[0]{*}{\begin{sideways}\rotatebox[origin=c]{270}{VGG19}\end{sideways}} 
& \multirow{2}[1]{*}{1\%} & Before & 1.00 & 0.631 & 0.825 & 1.00 \\
\cline{3-7}          &       & \cellcolor[rgb]{ .816,  .816,  .816}After & \cellcolor[rgb]{ .816,  .816,  .816}0.456 & \cellcolor[rgb]{ .816,  .816,  .816}0.111 & \cellcolor[rgb]{ .816,  .816,  .816}0.6 & \cellcolor[rgb]{ .816,  .816,  .816}0.24 \\
\cline{2-7}          & \multirow{2}[1]{*}{10\%} & Before & 1.00 & 0.837 & 0.835 & 1.00 \\
\cline{3-7}          &       & \cellcolor[rgb]{ .816,  .816,  .816}After & \cellcolor[rgb]{ .816,  .816,  .816}0.599 & \cellcolor[rgb]{ .816,  .816,  .816}0.267 & \cellcolor[rgb]{ .816,  .816,  .816}0.528 & \cellcolor[rgb]{ .816,  .816,  .816}0.501 \\
\cline{2-7}          & \multirow{2}[1]{*}{25\%} & Before & 1.00 & 0.857 & 0.828 & 1.00 \\
\cline{3-7}          &       & \cellcolor[rgb]{ .816,  .816,  .816}After & \cellcolor[rgb]{ .816,  .816,  .816}0.334 & \cellcolor[rgb]{ .816,  .816,  .816}0.211 & \cellcolor[rgb]{ .816,  .816,  .816}0.604 & \cellcolor[rgb]{ .816,  .816,  .816}0.228 \\
\cline{2-7}          & \multirow{2}[1]{*}{50\%} & Before & 0.997 & 0.877 & 0.81 & 1.00 \\
\cline{3-7}          &       & \cellcolor[rgb]{ .816,  .816,  .816}After & \cellcolor[rgb]{ .816,  .816,  .816}0.765 & \cellcolor[rgb]{ .816,  .816,  .816}0.208 & \cellcolor[rgb]{ .816,  .816,  .816}0.619 & \cellcolor[rgb]{ .816,  .816,  .816}0.135 \\
\hline
\end{tabular}%
\label{tab:unlearn_MIA_acc}%
\vspace{-2mm}
\end{table}%


Table~\ref{tab:unlearn_MIA_acc} demonstrates the effectiveness of our unlearning algorithm based on the the white-box MIA success against forgetting data. Prior to unlearning, all forgetting data samples within $D_f$ received high probabilities from the MIA model, indicating that all data points in $D_f$ were likely used in training the target model. For instance, when considering forgetting data that includes 50\% of the training data samples, all target models from the four deep learning architectures initially exhibited nearly 100\% accuracy in membership fingerprinting across all four datasets.

\begin{figure*}[!th]
    \centering
    {\includegraphics[width=\textwidth]{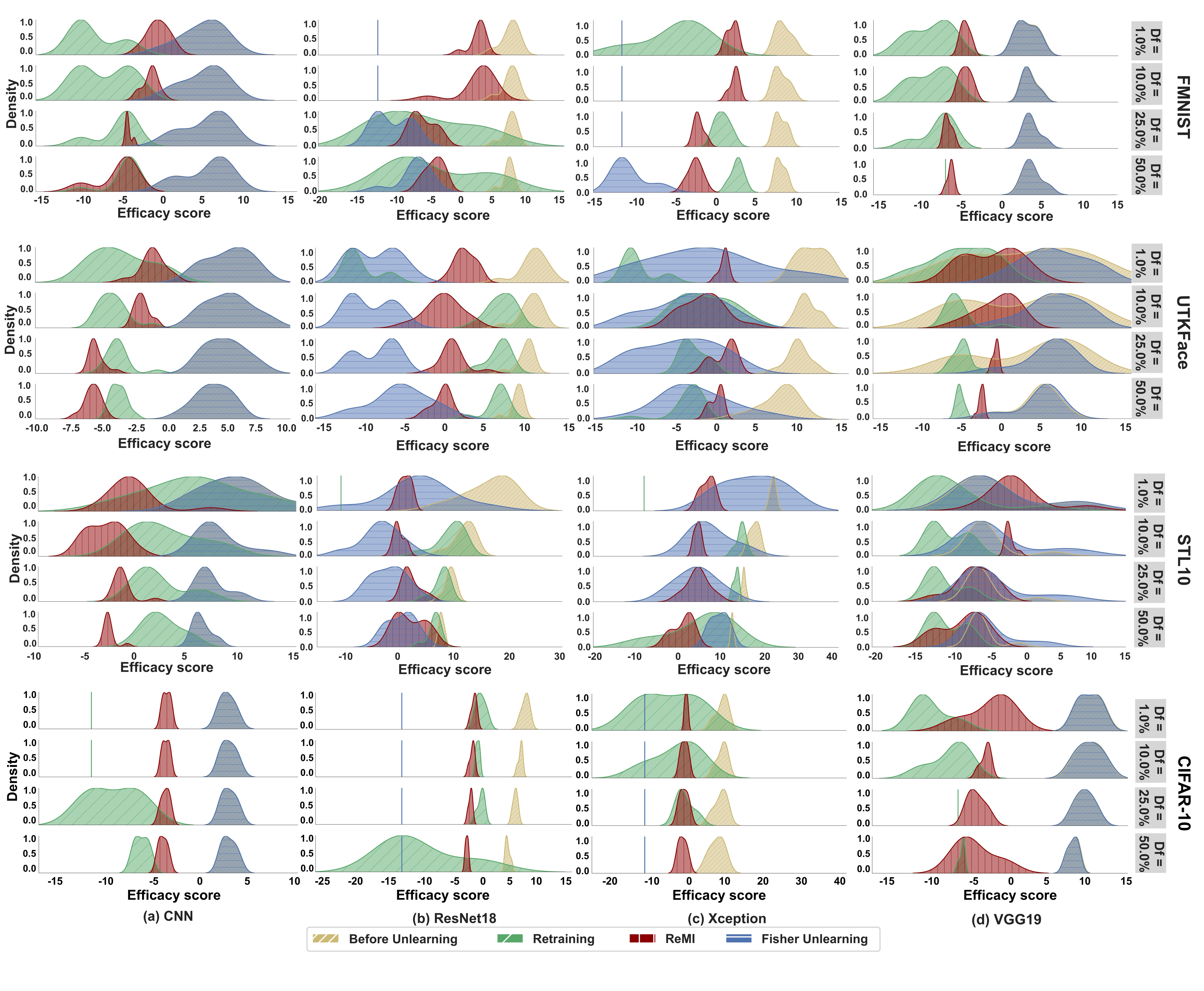}}
    \vspace{-12mm}
    \caption{Distribution of efficacy scores on forgetting data for the target models before and after unlearning across four datasets which shows how much information the model can leak. The results indicate that our unlearning method, ReMI, exposes less information compared to the original target model before unlearning. Additionally, the efficacy of the unlearning algorithms exhibits variations depending on the complexity of the deep learning architecture employed in the target model and the size of the forgetting dataset.}
    \label{fig:fnr_fpr_cdf}
    \vspace{-4mm}
\end{figure*}

Applying \sol, however, significantly reduces the white-box MIA accuracy on all four target models across all four datasets. Specifically, after applying the unlearning process to the target models trained on the CIFAR-10 dataset, the MIA accuracy on the forgetting data ($D_f|_{0.5}$) for each of the four deep learning methods is as follows: 4.4\% (CNN), 0.1\% (ResNet18), 0.0\% (Xception), and 13.5\% (VGG19). Similar unlearning performance is observed on the STL10, UTKFace, and FMNIST datasets. For example, on the FMNIST dataset, the attack accuracy decreases from the range 88\%--98\% for the original target models to a range of 3.9\% for the Xception model and 0.0\% for the ResNet18 model. On the STL10 dataset, the attack accuracy drops from the range 81\%--100\% to the range 3.1\%--10\% across all architectures except VGG19. 
Regarding VGG19, it did not perform as well as other architectures in unlearning the samples. We attribute this behavior to the lack of skip connections in the VGG architecture, which results in highly non-smooth loss surfaces~\cite{li2018visualizing}. To the best of our knowledge, no other study has used the VGG architecture for unlearning, leaving no baseline for comparison.
The results in Table~\ref{tab:unlearn_target_acc} and Table~\ref{tab:unlearn_MIA_acc} highlight the effectiveness of our unlearning algorithm in eliminating the private information of forgetting data from the target models without compromising the accuracy of the target classification models. We also evaluated black-box MIAs on the unlearned data and included the results in the supplementary material due to space limitation.
\begin{figure*}[!]
    \centering
    {\includegraphics[width=\textwidth]{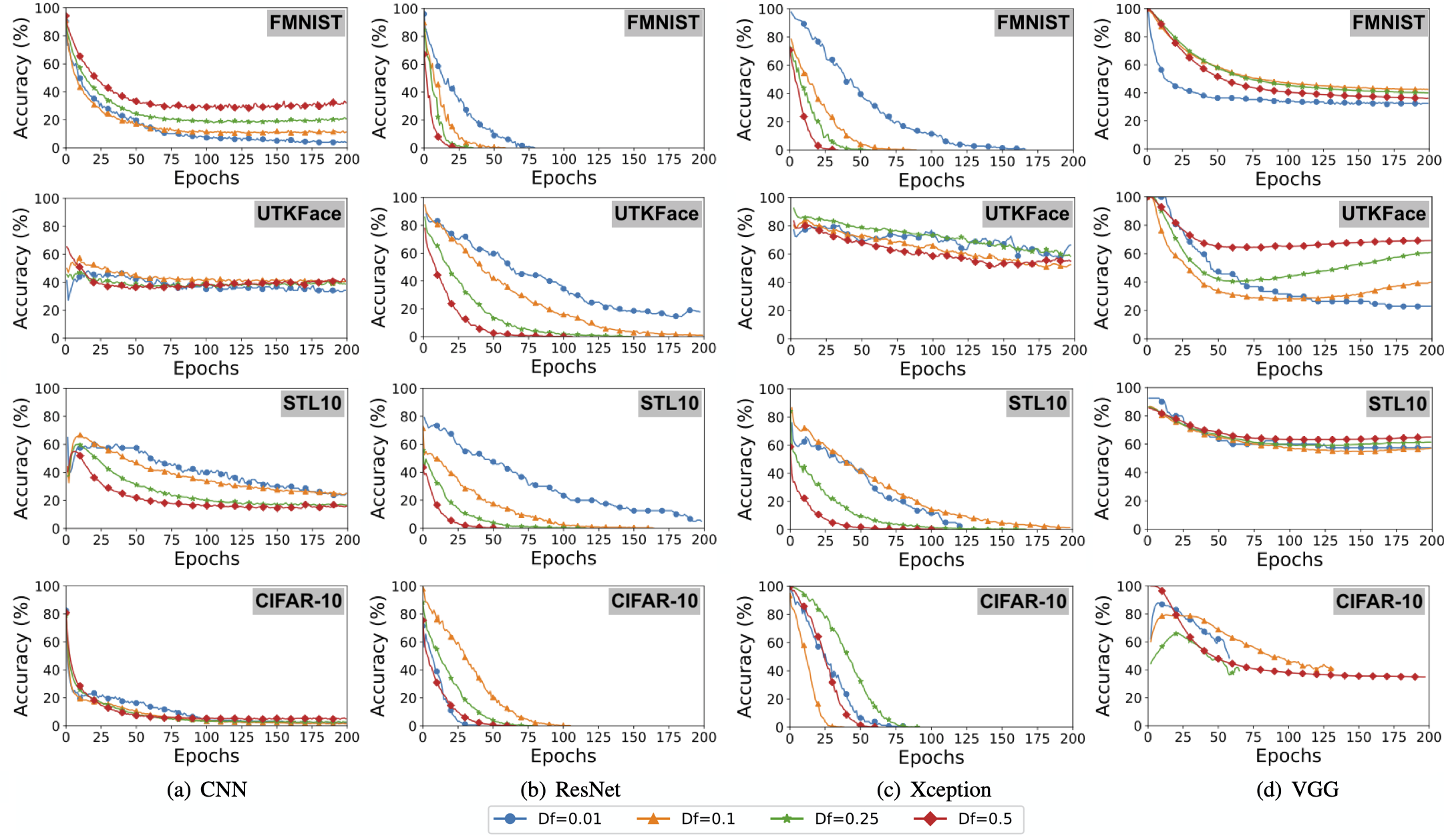}}
    \vspace{-7mm}
    \caption{Comparison of white-box MF accuracy during unlearning, highlighting the importance of the forgetting data size and the model's architecture on unlearning convergence. Selecting a smaller data to forget leads to faster convergence, particularly on CNN and VGG19.} 
    \label{fig:MIA_over_epoch}
    \vspace{-4mm}
\end{figure*}
%

Besides classification fidelity of the unlearned model, the unlearning efficacy score is another valuable metric for assessing the effectiveness of machine unlearning techniques~\cite{BecLie22,c3}. The efficacy score quantifies the degree to which a 
model retains or discloses information by accounting principles from information theory and epistemic uncertainty~\cite{BecLie22}. A more robust unlearning algorithm is expected to lead to an unlearned model with a lower efficacy score, indicating higher uncertainty about the target data. We note that we calculate the efficacy score only for the forget data.
Moreover, we conducted unlearning experiments using the same model and data configurations with different random seeds to collect multiple efficacy scores for distribution analysis. This allows us to compare the efficacy score of \sol against other unlearning algorithms.

We evaluate the effectiveness of \sol by comparing its efficacy score with that of two alternative methods: Fisher Unlearning and Naive Retraining.  The experiments for Fisher Unlearning and Naive Retraining were conducted under identical configurations as those used for our unlearning algorithm.  These configurations included using four deep learning architectures and forgetting data of varying sizes (1.00\%, 10.0\%, 25.0\%, and 50.0\%) from four datasets. Figure~\ref{fig:fnr_fpr_cdf} presents the distribution of efficacy scores obtained from the original target model and the unlearned models resulting from all three unlearning algorithms. The results indicate that 
in most cases, the three unlearning algorithms produced new models with lower efficacy scores than the original target model. This trend was consistent across various deep learning architectures regardless of the size of the forgetting data. Notably, while the efficacy distribution of Fisher Unlearning sometimes overlapped with the efficacy distribution of the original model, as observed in cases like CNN on all four datasets and VGG19 on the FMNIST \& CIFAR-10 datasets, the overall trend indicated a decrease in model efficacy.
%

%
%
%

Furthermore, the efficacy of the unlearning algorithms exhibited variations depending on the deep learning architecture employed in the target model. For example, both our algorithm ReMI and Model Retraining generated new models with significantly lower efficacy scores across all four forgetting datasets when the target model was based on CNN architecture. Our algorithm, in particular, demonstrated superior efficacy distributions for forgetting data sizes of 25\% and 50\%. However, Fisher unlearning outperformed the others by producing unlearning models with lower efficacy scores for larger forgetting datasets, specifically those with sizes of 10.0\%, 25.0\%, and 50.0\%, when the target model is ResNet18. Nevertheless, our unlearning algorithm still exhibited better unlearning efficacy than Model Retraining across various settings.

This analysis underscores the significant influence of both model complexity and forgetting dataset size on the performance of Model Retraining and Fisher Unlearning. 
For example, Model Retraining tends to yield less effective unlearning models as the model complexity increases, transitioning from simpler models like CNN to moderately complex models like ResNet. In contrast, Fisher Unlearning excels in generating better models with lower efficacy scores as the size of the forgetting dataset grows. Our unlearning algorithm, positioned between these two methods, is less affected by variations in model complexity and forgetting data sizes, offering a balanced performance across a range of scenarios.

\begin{figure*}[!th]
    \centering
    {\includegraphics[width=\textwidth]{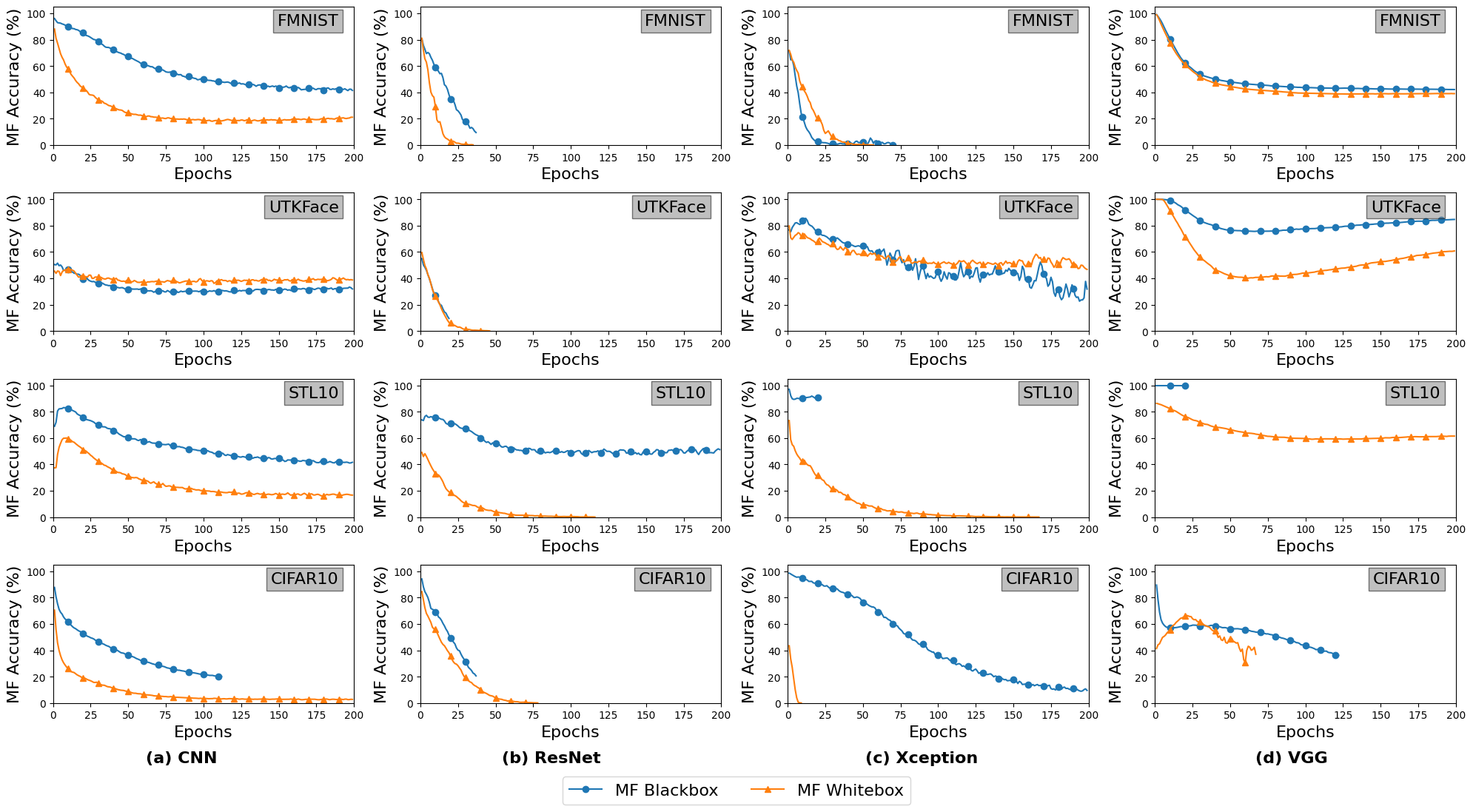}}
    \vspace{-7mm}
    \caption{MF accuracy during unlearning versus epochs for black-box and white-box settings for $D_f$ is 0.25. The results suggest using white-box MF to guide the unlearning process will have faster convergence, which consequently speeds up the unlearning process.}
    \label{fig:MF-whitebox-blackbox}
    \vspace{-3mm}
\end{figure*}

\noindent
\textbf{Unlearning Process Discussion}: Figure~\ref{fig:MIA_over_epoch} provides insights into how the accuracy of the membership fingerprinting (MF) model on the forgetting dataset changes over various training epochs. The results consistently demonstrate a decrease in the MF accuracy during the unlearning process across different target models and forgetting data sizes. 
The speed of convergence depends on both the forgetting data size and the specific deep learning models used. For instance, ResNet and Xception target models across all datasets exhibit faster convergence with bigger forgetting data sizes (25.0\% and 50.0\%), while smaller forgetting datasets (1.0\% and 10.0\%) require more epochs for effective convergence. 
A different pattern is observed with CNN and VGG19 models across all datasets, where the unlearning process was longer despite the forget dataset.

Figure~\ref{fig:MF-whitebox-blackbox} provides an insightful representation of how the MF accuracy on the forgetting dataset changes across different training epochs when employing different types of MF (black-box and white-box) as privacy approximation functions ($\mathcal{G}(I(w))$) during the unlearning process when the forgetting data size is 25\%. The other forgetting data size results can be found in the supplementary document.
The analysis illustrates that using the white-box MF model as a reference for minimizing the distribution of attack probabilities between in-sample and out-of-sample data proves to be more effective than relying on the black-box MF model. This approach results in a faster convergence of the unlearning loss and lower data leakage for the forgetting data when compared to the utilization of the black-box model. This observation can be partially attributed to the set of white-box features, which include gradients derived from the target model, contributing to enhanced unlearning performance.
On the other hand, employing the black-box MF model to guide the unlearning process demonstrates less success, particularly evident when applying CNN and ResNet18 on STL10 data. This analysis underscores the need for more effective privacy approximation functions to enhance unlearning performance.

\noindent
\textbf{Cross-attack Evaluation}: We further assess the impact of our algorithm on privacy information exposure under different membership fingerprinting scenarios. Specifically, we evaluate the accuracy of a black-box MF 
against models unlearned using a white-box MF as a reference, and vice versa. This bidirectional evaluation aims to uncover insights into the effectiveness of our algorithm in mitigating privacy risks and understanding how different fingerprinting strategies interact with one another. 
Table~\ref{tab:unlearn_attack} presents a summary of the results from our comprehensive bidirectional membership fingerprinting evaluations across various forgetting dataset sizes. The successful unlearning process, guided by a reference MF model, should effectively mitigate privacy risks across different mechanisms, indicating the removal of private information associated with the training data.  
As presented in Table~\ref{tab:unlearn_attack}, when the target model undergoes unlearning by incorporating white-box MF probabilities into the loss function optimization, the resulting new model exhibits significantly reduced attack accuracy when subjected to independent black-box fingerprinting. This observation holds true across unlearned models trained with forgetting data of varying sizes. For instance, the fingerprinting accuracy of the black-box MF diminishes from 100\% to 40\%, 62.5\%, 42.5\%, and 40\% when applied to forgetting datasets with sizes of 1\%, 10\%, 25\%, and 50\%, respectively, after the unlearning is applied based on white-box MF attack. In contrast, the target models that have undergone unlearning using a black-box MF model remain vulnerable to white-box MF.

\begin{table}[t]
  \centering
  \caption{Cross-method Unlearning Efficacy Evaluation}
    \vspace{-3mm}
    \begin{tabular}{|c|c|c|c|c|}
    \hline
    \multirow{2}[0]{*}{$D_f$} & \multirow{2}[0]{*}{Unlearning Source} & \multirow{2}[0]{*}{Evaluation} & \multicolumn{2}{c|}{Attack Accuracy (\%)} \\
\cline{4-5}          &       &       & Before  & After \\
    \hline
    \multirow{2}[0]{*}{0.01} & White-box & Black-box & {\bf 100}   & {\bf 40} \\
\cline{2-5}          & Black-box & White-box & 92.5  & 87.5 \\
    \hline
    \multirow{2}[0]{*}{0.1} & White-box & Black-box & {\bf 100}   & {\bf 62.5} \\
\cline{2-5}          & Black-box & White-box & 92.5  & 70 \\
    \hline
    \multirow{2}[0]{*}{0.25} & White-box & Black-box & {\bf 100}   & {\bf 42.5} \\
\cline{2-5}          & Black-box & White-box & 92.5  & 100 \\
    \hline
    \multirow{2}[0]{*}{0.5} & White-box & Black-box & {\bf 100}   & {\bf 40} \\
\cline{2-5}          & Black-box & White-box & 92.5  & 92.5 \\
    \hline
    \end{tabular}%
  \label{tab:unlearn_attack}%
  \vspace{-2mm}
\end{table}%


\begin{table}[tbp]
  \centering
  \footnotesize
  \caption{Unlearning latency comparison of Retraining and ReMI Unlearning on UTKFace dataset. Time is measured in seconds. ReMI Unlearning includes total unlearning time and time for attack Loss calculation. The results include \sol's speed up over retraining.}
  \vspace{-0mm}
    \begin{tabular}{|c|c|c|c|c|c|}
    \hline
     \multirow{2}[0]{*}{}  & \multirow{2}[0]{*}{$D_f$} & \multirow{2}[0]{*}{} & \multicolumn{2}{c|}{ReMI Unlearning} & \multirow{2}[5]{*}{\shortstack{Speed \\ up} } \\
\cline{4-5}          &     Ratio  &  \shortstack{Retraining\\{\color{white}a}}   & \shortstack{Unlearning by\\ Eqn. \ref{eqn:losstol}}  & \shortstack{$\mathcal{L}_{\mathcal{G}} (w)$ \\ calculation} &  \\
    \hline
    \multirow{4}[0]{*}{\rotatebox[origin=c]{90}{CNN}} & 0.01  & 812.321 & 194.523 & 99.99 & 4.1x \\
\cline{2-6}          & 0.1   & 850.325 & 270.449 & 215.573 & 3.1x \\
\cline{2-6}          & 0.25  & 864.031 & 68.613 & 57.570 & 12.5x  \\
\cline{2-6}          & 0.5   & 808.766 & 204.579 & 176.096 & 3.9x  \\
    \hline
    \multirow{4}[0]{*}{\rotatebox[origin=c]{90}{ResNet}} & 0.01  & 1063.228 & 11.463 & 7.896 & 92.76x  \\
\cline{2-6}          & 0.1   & 1034.444 & 12.555 & 9.127 & 82.3x  \\
\cline{2-6}          & 0.25  & 950.442 & 163.488 & 123.163 & 5.81x  \\
\cline{2-6}          & 0.5   & 943.834 & 168.435 & 130.271 & 5.6x  \\
    \hline
    \multirow{4}[0]{*}{\rotatebox[origin=c]{90}{Xception}} & 0.01  & 1060.723 & 119.905 & 85.543 & 8.84x  \\
\cline{2-6}          & 0.1   & 787.714 & 312.653 & 231.046 & 2.51x  \\
\cline{2-6}          & 0.25  & 995.286 & 1123.721 & 849.638 & 0.88x  \\
\cline{2-6}          & 0.5   & 930.395 & 468.84 & 358.547 & 1.98x  \\
    \hline
    \multirow{4}[0]{*}{\rotatebox[origin=c]{90}{VGG19}} & 0.01  & 1044.088 & 153.981 & 106.279 & 6.78x  \\
\cline{2-6}          & 0.1   & 1051.894 & 100.856 & 70.592 & 10.42x  \\
\cline{2-6}          & 0.25  & 912.485 & 375.815 & 269.564 & 2.42x  \\
\cline{2-6}          & 0.5   & 833.932 & 540.613 & 351.665 &  1.54x \\
    \hline
\multicolumn{2}{|c|}{Avg. Time} & 933.994 & 268.155 & 196.410 & 3.48x  \\
    \hline
    \end{tabular}%
  \label{tab:unlearning_latency}%
  \vspace{-2mm}
\end{table}%

\noindent
\textbf{Unlearning Latency Evaluation}: Finally, we conducted an evaluation of the unlearning latency of our algorithm in comparison to Model Retraining. Table~\ref{tab:unlearning_latency} presents the results of our latency analysis based on the experiments we conducted. The evaluation involves measuring the running time of the unlearning process, which begins with the loading of a pre-trained target model and the specified forgetting data and ends when the unlearning process is completed. As demonstrated in Table~\ref{tab:unlearning_latency}, the ReMI unlearning algorithm exhibits an average completion time of approximately 268 seconds across all architectures for the UTKFace dataset. This is nearly four times faster than Model Retraining, which requires approximately 933 seconds on average. Notably, the most time-consuming step in the ReMI unlearning process is the calculation of attack probabilities for loss optimization ($\mathcal{L}_{\mathcal{G}} (w)$), which consumes approximately 196 seconds on average. This analysis underscores the lower latency of our unlearning algorithm. We did not compare with Fisher unlearning as it has shown to be an order of magnitude slower than retraining~\cite{FosSchBri24}.


%
%
\section{Conclusion}
We introduced \sol~-- a framework for removing samples of training data and their impact from trained neural networks. At its core, \sol can use various privacy approximation functions, which measure the information leakage of the model on forgotten data, to guide the unlearning process. In particular, we used MIA and membership fingerprinting models as our approximation functions. We designed a novel unlearning loss function to integrate the target classification loss and membership inference loss, ensuring the unlearned model achieves high unlearning efficacy and classification accuracy. Our empirical results, coupled with theoretical upper-bound analysis through a membership inference mechanism, showed the superiority of our proposed unlearning mechanism. 
%
\balance
\bibliographystyle{IEEEtran}
\bibliography{paper/ref}

\appendix
\section{Experiment Configuration and Supplementary Results}
\label{apn} 
\begin{table*}[!ht]
  \centering
  \small
  \caption{Hyperparameters of Unlearning algorithm}
    \vspace{-2mm}
    \begin{tabular}{|c|c|c|c|c|c|c|}
    \hline
    \multirow{2}[0]{*}{\textbf{Dataset}} & \multirow{2}[0]{*}{\textbf{Architecture}} & \multicolumn{4}{c|}{\textbf{Learning rate $\eta$}} & \multirow{2}[4]{*}{\textbf{$\lambda_2$}} \\
\cline{3-6}          &       & \textbf{$|D_f|=0.01$ } & \textbf{$|D_f|=0.1$ } & \textbf{$|D_f|=0.25$} & \textbf{$|D_f|=0.5$ } &  \\
    \hline
    \multirow{4}[0]{*}{\rotatebox[origin=c]{0}{FMNIST}} & CNN   & 0.01  & 0.01  & 0.01  & 0.01  & 0.98 \\
\cline{2-7}          & ResNet18 & 0.01  & 0.01  & 0.01  & 0.01  & 0.98 \\
\cline{2-7}          & Xception & 0.01  & 0.01  & 0.01  & 0.01  & 0.98 \\
\cline{2-7}          & VGG19 & 0.001 & 0.001 & 0.001 & 0.001 & 0.98 \\
    \hline
    \multirow{4}[0]{*}{\rotatebox[origin=c]{0}{UTKFace}} & CNN   & 0.1   & 0.1   & 0.1   & 0.1   & 0.98 \\
\cline{2-7}          & ResNet18 & 0.1   & 0.1   & 0.1   & 0.1   & 0.98 \\
\cline{2-7}          & Xception & 0.1   & 0.1   & 0.1   & 0.1   & 0.98 \\
\cline{2-7}          & VGG19 & 0.001 & 0.001 & 0.001 & 0.001 & 0.98 \\
    \hline
    \multirow{4}[0]{*}{\rotatebox[origin=c]{0}{STL10}}& CNN   & 0.1   & 0.1   & 0.1   & 0.1   & 0.98 \\
\cline{2-7}          & ResNet18 & 0.1   & 0.1   & 0.1   & 0.1   & 0.98 \\
\cline{2-7}          & Xception & 0.1   & 0.05  & 0.05  & 0.05  & 0.98 \\
\cline{2-7}          & VGG19 & 0.0005 & 0.0005 & 0.0005 & 0.0005 & 0.98 \\
    \hline
    \multirow{4}[0]{*}{\rotatebox[origin=c]{0}{CIFAR10}} & CNN   & 0.01  & 0.01  & 0.01  & 0.01  & 0.98 \\
\cline{2-7}          & ResNet18 & 0.01  & 0.01  & 0.01  & 0.01  & 0.98 \\
\cline{2-7}          & Xception & 0.05  & 0.05  & 0.01  & 0.01  & 0.98 \\
\cline{2-7}          & VGG19 & 0.1   & 0.05  & 0.05  & 0.001 & 0.98 \\
    \hline
    \end{tabular}%
  \label{tab:hyperparameter_unlearning}%
    \vspace{-3mm}
\end{table*}%

\begin{figure*}[!ht]
    \centering
    {\includegraphics[width=\textwidth]{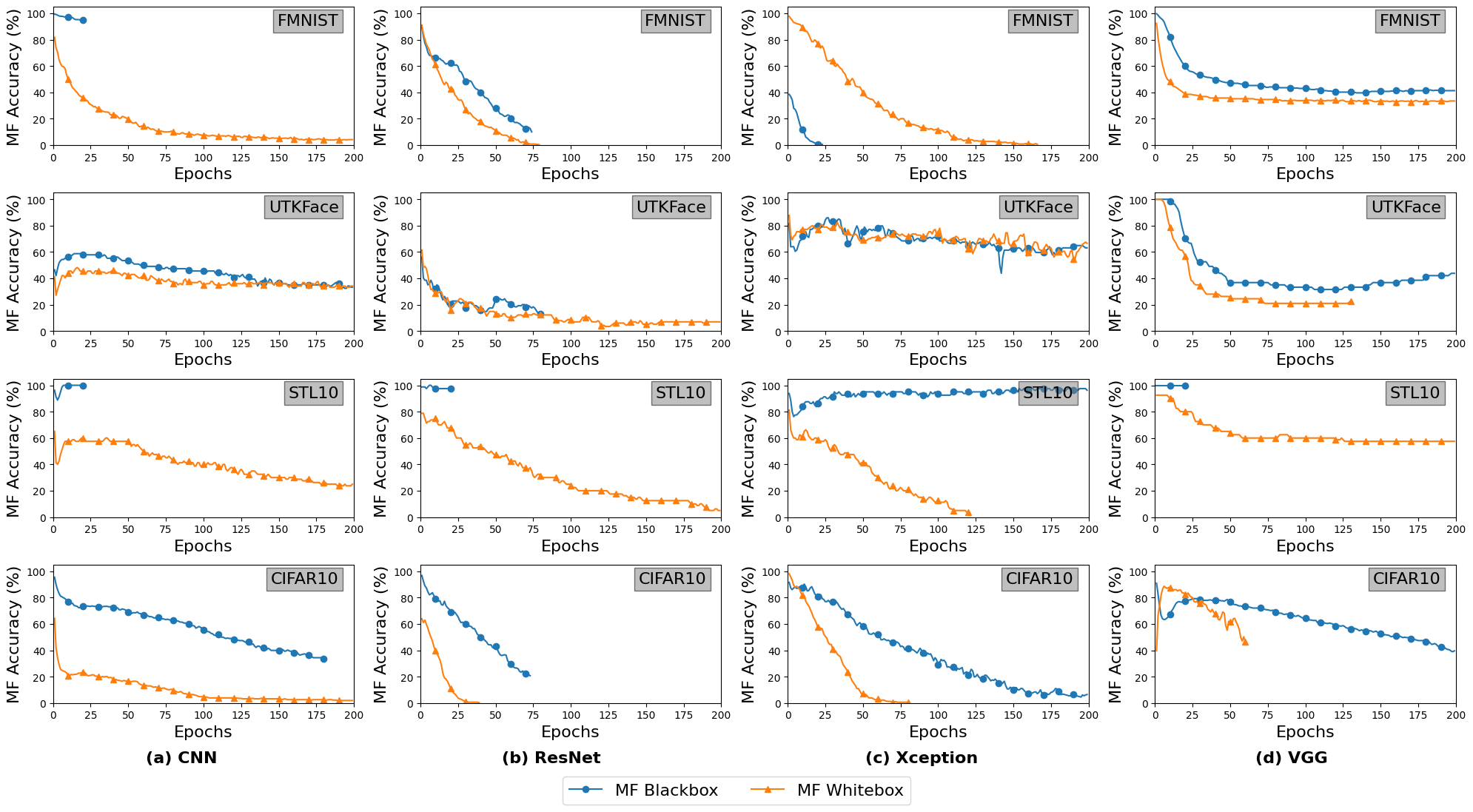}}
    \vspace{-7mm}
    \caption{MF accuracy during unlearning versus epochs for black-box and white-box settings for $D_f$ is 0.01. The results suggest using white-box MF to guide the unlearning process will have faster convergence, which consequently speeds up the unlearning process.}
    \label{fig:MF-whitebox-blackbox_1}
\end{figure*}

\begin{figure*}[!th]
    \centering
    {\includegraphics[width=\textwidth]{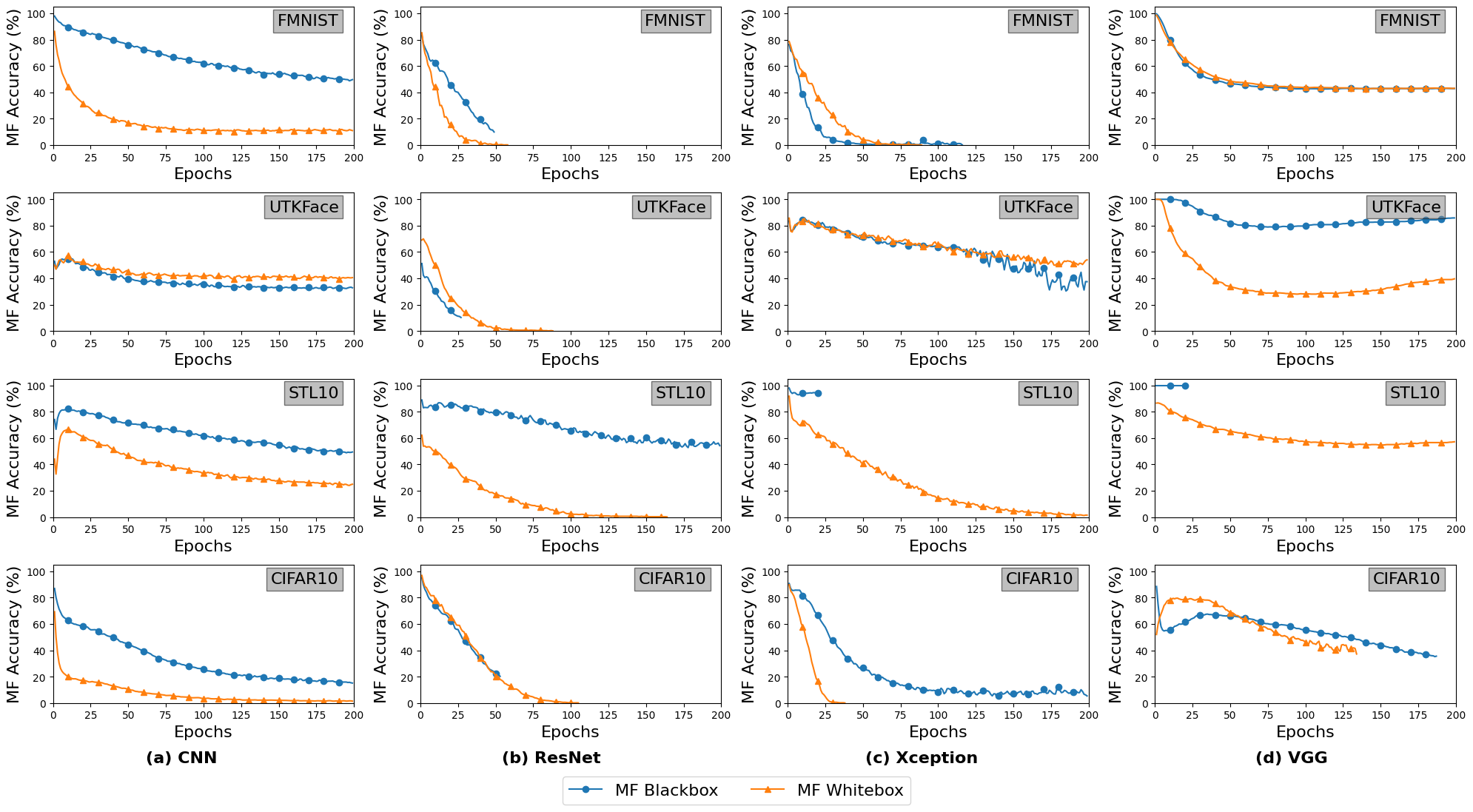}}
    \vspace{-7mm}
    \caption{MF accuracy during unlearning versus epochs for black-box and white-box settings for $D_f$ is 0.1. The results suggest using white-box MF to guide the unlearning process will have faster convergence, which consequently speeds up the unlearning process.}
    \label{fig:MF-whitebox-blackbox_10}
\end{figure*}

\begin{figure*}[!th]
    \centering
    {\includegraphics[width=\textwidth]{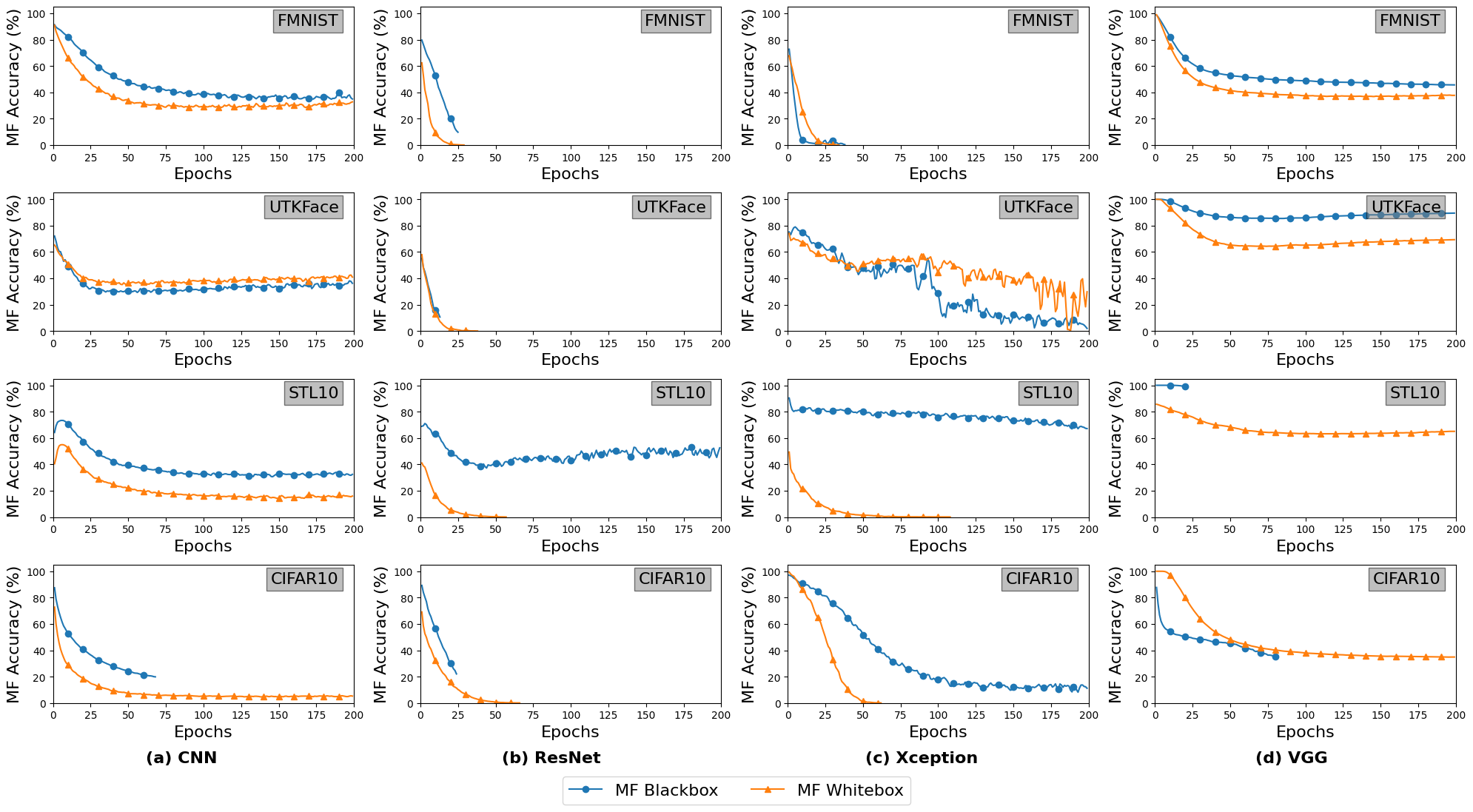}}
    \vspace{-7mm}
    \caption{MF accuracy during unlearning versus epochs for black-box and white-box settings for $D_f$ is 0.5. The results suggest using white-box MF to guide the unlearning process will have faster convergence, which consequently speeds up the unlearning process.}
    \label{fig:MF-whitebox-blackbox_50}
\end{figure*}

\clearpage

\begin{figure}[H]
    \centering
    \includegraphics[width=\columnwidth]{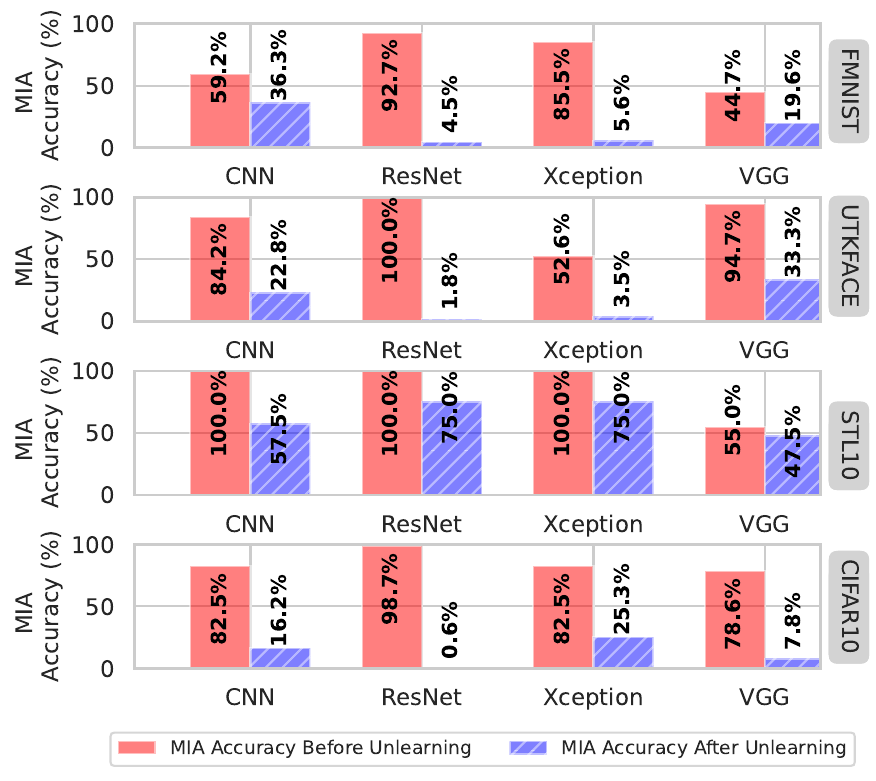}
    \vspace{-0mm}
    \caption{Blackbox MIA accuracy on the forget dataset, a subset of the target model's training dataset, before and after unlearning when the ratio of $D_f$ is 0.01. It is evident that \sol unlearning effectively reduces the information leakage of the forgetting data.}
    \label{fig:attack_accuracy_unlearning_001}
    \vspace{-0mm}
\end{figure}

\begin{figure}[h]
    \centering
    \includegraphics[width=\columnwidth]{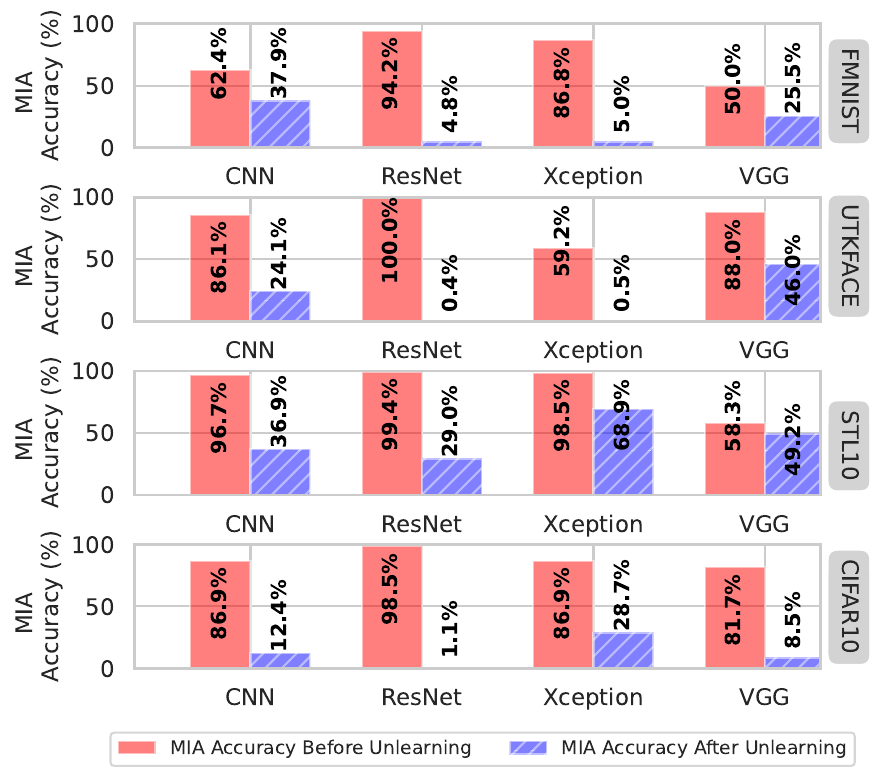}
    \vspace{-0mm}
    \caption{Blackbox MIA accuracy on the forget dataset, a subset of the target model's training dataset, before and after unlearning when the ratio of $D_f$ is 0.1. It is evident that \sol unlearning effectively reduces the information leakage of the forgetting data.}
    \label{fig:attack_accuracy_unlearning_01}
    \vspace{-0mm}
\end{figure}

\begin{figure}[h]
    \centering
    \includegraphics[width=\columnwidth]{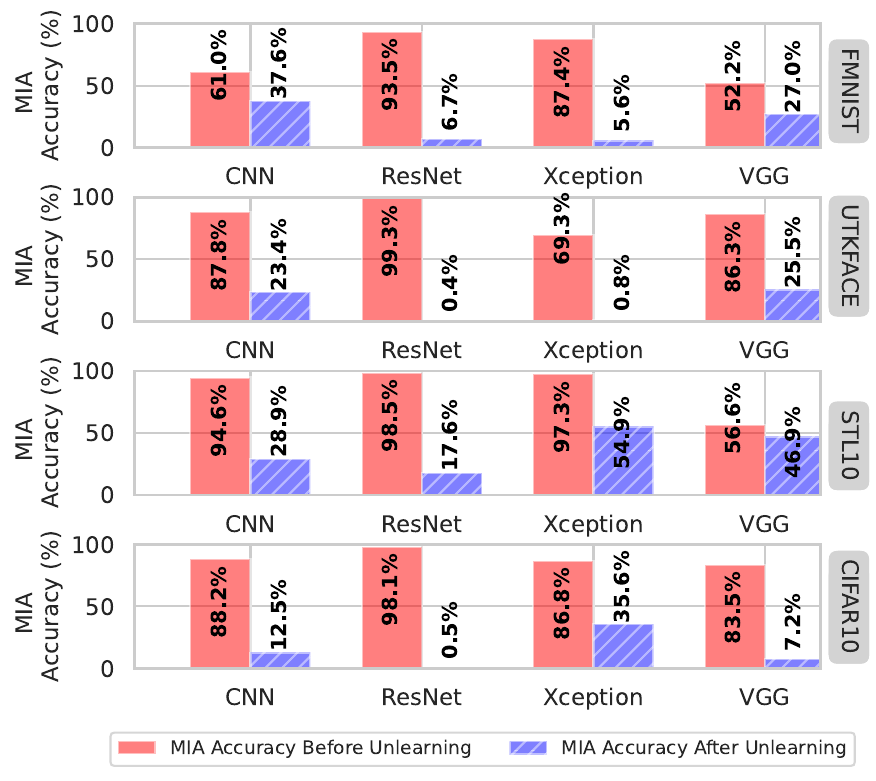}
    \vspace{-0mm}
    \caption{Blackbox MIA accuracy on the forget dataset, a subset of the target model's training dataset, before and after unlearning when the ratio of $D_f$ is 0.25. It is evident that \sol unlearning effectively reduces the information leakage of the forgetting data.}
    \label{fig:attack_accuracy_unlearning_025}
    \vspace{-0mm}
\end{figure}

\begin{figure}[H]
    \centering
    \includegraphics[width=\columnwidth]{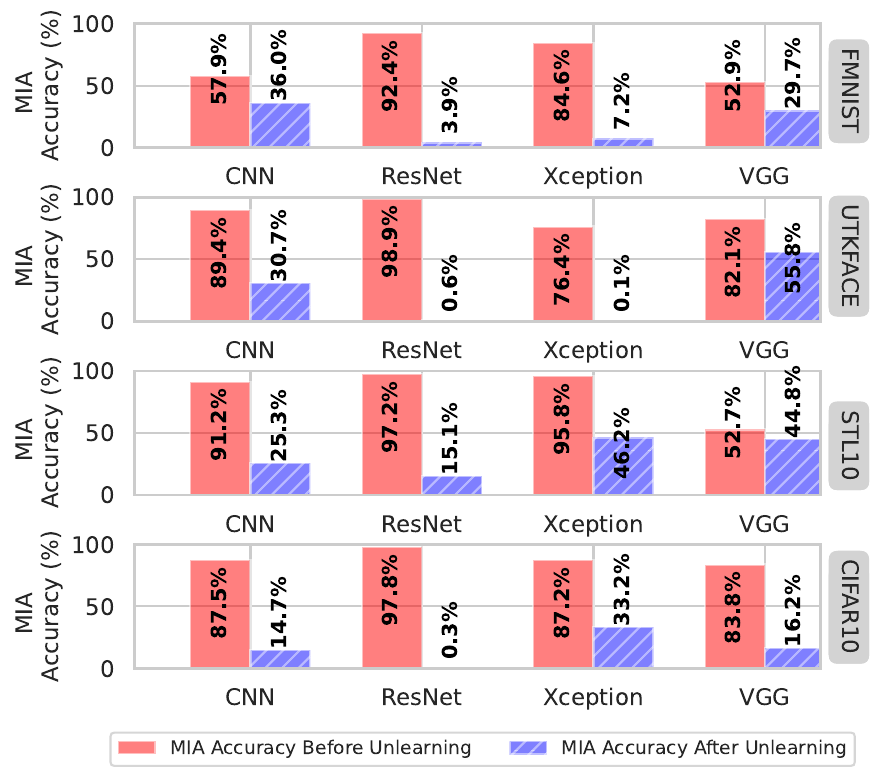}
    \vspace{-0mm}
    \caption{Blackbox MIA accuracy on the forget dataset, a subset of the target model's training dataset, before and after unlearning when the ratio of $D_f$ is 0.5. It is evident that \sol unlearning effectively reduces the information leakage of the forgetting data.}
    \label{fig:attack_accuracy_unlearning_05}
    \vspace{-0mm}
\end{figure}

\end{document}